\documentclass[final,3p,times,twocolumn,authoryear]{elsarticle}

\usepackage[labelfont=bf]{caption}
\usepackage[hidelinks]{hyperref}
\hypersetup{
    colorlinks = True
    }
\usepackage{colortbl}
\usepackage{float}
\usepackage{supertabular}
\usepackage{rotating}
\usepackage{booktabs}
\usepackage{ragged2e}
\usepackage[longtable]{multirow}
\usepackage{longtable}
\usepackage{array}
\usepackage{graphicx}
\usepackage{verbatim} %comments
\usepackage{apalike}
\usepackage{interval}
\usepackage{amsmath,amssymb,amsfonts}
\usepackage{algorithmic}
\usepackage{adjustbox}
\usepackage{textcomp}
\usepackage{multicol}
\usepackage{apalike}
\usepackage{subcaption}

\journal{Decision Analytics Journal}

\begin{document}
\begin{frontmatter}

\title{A Natural Language Processing-Based Classification and Mode-Based Ranking of Musculoskeletal Disorder Risk Factors}

\author[label1]{Md Abrar Jahin \corref{cor1}}
\ead{abrar.jahin.2652@gmail.com}

\author[label1]{Subrata Talapatra}
\ead{subrata@iem.kuet.ac.bd}

\cortext[cor1]{Corresponding author.}
\address[label1]{Department of Industrial Engineering and Management, Khulna University of Engineering \& Technology (KUET), Khulna, 9203, Bangladesh}

%\cortext[contrib]{These authors contributed equally to this work and share the first authorship.}

\begin{abstract}
This research explores the intricate landscape of Musculoskeletal Disorder (MSD) risk factors, employing a novel fusion of Natural Language Processing (NLP) techniques and mode-based ranking methodologies. Enhancing knowledge of MSD risk factors, their classification, and their relative severity is the main goal of enabling more focused preventative and treatment efforts. The study benchmarks eight NLP models, integrating pre-trained transformers, cosine similarity, and various distance metrics to categorize risk factors into personal, biomechanical, workplace, psychological, and organizational classes. Key findings reveal that the Bidirectional Encoder Representations from Transformers (BERT) model with cosine similarity attains an overall accuracy of 28\%, while the sentence transformer, coupled with Euclidean, Bray-Curtis, and Minkowski distances, achieves a flawless accuracy score of 100\%. Using a 10-fold cross-validation strategy and performing rigorous statistical paired t-tests and Cohen's d tests (with a 5\% significance level assumed), the study provides the results with greater validity. To determine the severity hierarchy of MSD risk variables, the research uses survey data and a mode-based ranking technique parallel to the classification efforts. Intriguingly, the rankings align precisely with the previous literature, reaffirming the consistency and reliability of the approach. ``Working posture" emerges as the most severe risk factor, emphasizing the critical role of proper posture in preventing MSD. The collective perceptions of survey participants underscore the significance of factors like ``Job insecurity," ``Effort reward imbalance," and ``Poor employee facility" in contributing to MSD risks. The convergence of rankings provides actionable insights for organizations aiming to reduce the prevalence of MSD. The study concludes with implications for targeted interventions, recommendations for improving workplace conditions, and avenues for future research. This holistic approach, integrating NLP and mode-based ranking, contributes to a more sophisticated comprehension of MSD risk factors and opens the door for more effective strategies in occupational health.

\end{abstract}

\begin{keyword}
Musculoskeletal Disorder (MSD) \sep Risk Factors \sep Natural Language Processing (NLP) \sep Occupational Health and Safety \sep Machine Learning
\end{keyword}

\end{frontmatter}

\section{Introduction}
\label{sec:introduction}
The unexplored potential of Natural Language Processing (NLP) provides an intriguing frontier in occupational health, safety, and ergonomics. NLP's use in the field of musculoskeletal disorders (MSD) is primarily unexplored, despite the significant issues these disorders offer and the significant impact they have on both individuals and companies. By employing NLP to classify risk variables associated with MSD, this research aims to pave the way for developing more efficient preventive and treatment approaches. Effective preventive and management techniques for multiple sclerosis MSD rely heavily on identifying and classifying risk factors. Although these characteristics have been studied in previous research, a significant gap exists in classifying MSD risk factors based on artificial intelligence (AI). To close this gap, this study presents a novel method that uses eight different NLP models, each of which uses unique similarity or distance measurements. This research is necessary because it differs from other studies focusing on the importance of risk variables without a systematic classification. Furthermore, integrating NLP models, such as pre-trained transformers and diverse distance measures, yields a comprehensive study that presents quantitative and subtle linguistic viewpoints. By combining these two cutting-edge models, we want to close a significant gap in the MSD literature and advance the conversation on using AI approaches in advanced occupational health and safety research.

The human musculoskeletal system, encompassing muscles, tendons, nerves, bones, and joints, serves as a marvel of biology, facilitating movement and functionality (\citealp{leite_risk_2021}; \citealp{canetti_risk_2020}; \citealp{ghasemi_impacts_2020}; \citealp{bispo_risk_2022}). It allows individuals to partake in a diverse range of physical activities, yet it is susceptible to the strains and stresses of everyday life. Activities such as heavy lifting, repetitive motions, exposure to vibrations, and assuming awkward postures can lead to wear and tear, resulting in discomfort and injuries (\citealp{yang_types_2020}; \citealp{mondal_impact_2022}).

This gradual wear and tear culminate in a spectrum of conditions known as MSD, spanning a wide array of issues, from muscle strains and tendonitis to nerve-related conditions, joint disorders (e.g., gout, rheumatoid arthritis, osteoarthritis), bone problems (e.g., fractures, osteoporosis), spinal disk complications (e.g., degenerative disk, herniated disk), ligament injuries (e.g., ligament sprains), and circulatory challenges (\citealp{bayzid_prevalence_2019}; \citealp{bairwa_prevalence_2022}). MSDs do not manifest suddenly; rather, they develop gradually, often beginning with discomfort following physical exertion (\citealp{leite_risk_2021}). Failure to address these initial discomforts can lead to the accumulation of stress within the body, resulting in pain across various regions, including the upper and lower back, neck, shoulders, elbows, and extremities (e.g., forearms, legs, knees, feet, hands, fingers) (\citealp{whysall_stage_2006}). Inadequate treatment can exacerbate these conditions into acute musculoskeletal diseases (\citealp{rezaee_prevalence_2014}; \citealp{thetkathuek_work-related_2018}).

The roster of acute musculoskeletal diseases includes Carpal Tunnel Syndrome, Tendonitis, Tension Neck Syndrome, Epicondylitis, Radial Tunnel Syndrome, Hand Arm Vibration Syndrome, and Thoracic Outlet Compression, among others. These conditions can lead to permanent disabilities, significantly impacting organizational productivity (\citealp{woolf_bone_2000}; \citealp{bazazan_association_2019}). Consequently, preventing MSD problems has become paramount for individuals, organizations, and governments.

While MSDs are known to have low mortality rates, their impact on workforce functionality, healthcare expenditure, and absenteeism is substantial (\citealp{widanarko_interaction_2015}). In the United States, over 2 million employees grapple with MSD problems annually, with 30\% facing permanent disability (\citealp{silva_requirements_2019}). Similarly, in countries like Sweden and Canada, musculoskeletal injuries and diseases represent the leading causes of work-related absenteeism (\citealp{hartvigsen_what_2018}; \citealp{sakthi_nagaraj_evaluation_2019}). Furthermore, MSD overshadows organizational performance and leads to increased healthcare costs post-retirement. In Brazil, musculoskeletal issues accounted for 20\% of the nation's total sick pay and disability compensation between 2012 and 2016 (\citealp{dorsey_effectiveness_2016}).

The people of Bangladesh, like their global counterparts, frequently grapple with diverse MSD problems, affecting individuals of all ages, genders, and social classes (\citealp{bevan_economic_2015}; \citealp{bowers_psychological_2018}). Additionally, more than one-third of individuals in developing nations face MSD problems annually due to occupational factors (\citealp{summers_musculoskeletal_nodate}; \citealp{bispo_risk_2022}). Consequently, the financial repercussions of MSD problems extend beyond individual sufferers to encompass broader societal and organizational losses (\citealp{neupane_physical_2013}; \citealp{bowers_psychological_2018}).

In the extensive realm of research on MSD, a multitude of risk factors have been scrutinized. Researchers have diligently investigated the individual contributions of these factors in the development of MSD problems, aiming to unravel their intricate relationship with the occurrence of MSD (\citealp{mohseni-bandpei_risk_2006}; \citealp{baek_association_2018}). Indeed, the consensus among researchers is that this relationship is exceedingly complex (\citealp{kathy_cheng_work-related_2013}; \citealp{constantino_coledam_factors_2019}).

Consider, for instance, the influence of demographic factors. Studies have revealed that the susceptibility to MSD problems varies significantly with a person's age, gender, Body Mass Index (BMI), education, and work experience (\citealp{gallagher_examining_2013}; \citealp{tang_prevalence_2022}). These factors, often called personal risk factors, represent fundamental characteristics of an individual's demographic profile (\citealp{oha_individual_2014}; \citealp{jain_risk_2018}; \citealp{shohel_parvez_assessment_2022}). In parallel, the development of MSD problems is closely tied to physical factors encompassing working posture, repetitive motions, applied forces, deviations from neutral body alignment, vibrations, workspace layout, work pace, lighting conditions, environmental factors, and noise levels (\citealp{tang_prevalence_2022}). For instance, static or awkward working postures maintained over extended periods have been identified as significant physical risk factors contributing to the onset of MSD problems (\citealp{baek_association_2018}; \citealp{bairwa_prevalence_2022}). Furthermore, workplace conditions, such as exposure to high temperatures and humidity, have also been linked to a heightened risk of developing MSD problems (\citealp{leite_risk_2021}; \citealp{canetti_risk_2020}).

Beyond these tangible factors, the psychosocial aspects of a worker's environment also play a pivotal role. Factors related to human psychology and an organization's management abilities, such as job dissatisfaction, social support, mental and occupational stress, job insecurity, effort-reward imbalances, inadequate breaks, suboptimal job design, high job demands, management styles, and employee facilities, can induce mental or occupational stress among workers, potentially leading to MSD problems (\citealp{kolstrup_work-related_2012}; \citealp{bispo_risk_2022}). For example, job insecurity can trigger mental stress, which, in turn, may lead to depression—a condition that heightens the risk of developing MSD problems. Additionally, an organization's suboptimal job management policies can contribute to occupational and mental stress among employees, further elevating the likelihood of MSD problems.

In light of these intricate interplays, risk factors can be broadly categorized into two distinct groups based on their role in developing MSD problems—direct and indirect risk factors. Direct risk factors encompass elements that directly induce pain, injuries, disorders, or musculoskeletal diseases (\citealp{widanarko_interaction_2015}; \citealp{canetti_risk_2020}). Conversely, indirect risk factors are those that primarily lead to mental health issues, such as anxiety and depression, which subsequently increase the risk of MSD problems (\citealp{neupane_physical_2013}; \citealp{sakthi_nagaraj_evaluation_2019}; and Mathiyazhagan, 2019).

The significance and novelty of the research problem lie in its departure from traditional studies on MSD risk factors. While these studies have mostly focused on these factors, our study takes a bold step forward by implementing an NLP-based scientific classification and ranking framework. Past classification and ranking attempts have lacked AI implementation, hindering a comprehensive understanding of its potential in ergonomics. Furthermore, as shown by previous research (\citealp{thetkathuek_work-related_2018}; \citealp{bayzid_prevalence_2019}), not all MSD risk factors carry equal weight in causing problems. Hence, our key goals are to scientifically categorize these risk factors and analyze their specific contributions to the development of MSD. Because of the unsupervised nature of our approach, the computational requirement is optimum in our research regardless of the variation in dataset size.

The motivation behind this research is deeply rooted in the imperative need to advance the prevention and management of MSD problems. While prior research has offered valuable insights, an untapped opportunity exists to harness the capabilities of NLP in this domain. NLP has not been extensively adopted in ergonomics and occupational health research. This research seeks to bridge this gap and introduce NLP as a transformative tool for understanding and addressing MSD. NLP confronts a multitude of challenges, spanning language translation, biomedical applications, sentiment analysis, search engines, finance, product recommendations, and education. Despite its extensive reach, NLP has yet to make significant inroads into the domains of occupational health, safety, and ergonomics. Addressing this gap is the primary focus of the present article, marking a pioneering effort in utilizing AI to categorize MSD risk factors.

The selection of NLP and mode-based ranking methodologies in this article is deliberate and strategic. NLP is chosen for its inherent ability to decipher and categorize the nuanced language surrounding MSD risk factors, offering a systematic approach to extracting valuable insights from a vast body of literature. This technology enables the creation of a robust foundation for scientific risk factor classification. Concurrently, mode-based ranking methodologies are employed to discern the significance and prevalence of identified risk factors within the dataset, providing a quantitative lens to prioritize factors based on their frequency and impact. The synergy between NLP and mode-based ranking methodologies is evident in their complementary roles: NLP facilitates the qualitative understanding and classification of risk factors, while mode-based ranking adds a quantitative dimension, collectively enriching the research methodology and contributing to a more comprehensive analysis of MSD risk factors.

The following research contributions of our article collectively advance the understanding of MSD risk factors and provide actionable insights for effective prevention and management strategies:

\begin{itemize}
    \item Development of a novel framework that combines an unsupervised NLP-based approach with empirical data to categorize MSD risk factors into distinct classes.

    \item Implementation of eight distinct methods, including pre-trained transformers, similarity-based metrics (Cosine and Jaccard), and distance-based metrics (Euclidean, Manhattan, Minkowski, Mahalonobis, Bray-Curtis), to categorize risk factors with a high degree of accuracy.

    \item Identification of four approaches that achieved 100\% accuracy in categorizing 25 risk factors into five distinct classes, highlighting the effectiveness of the proposed methodology.

    \item Evaluation on the significance of the model's performance using statistical paired t-tests and Cohen's d tests.

    \item Integration of a large-scale survey involving 1050 participants to rank the severity of MSD risk factors on a scale of 1 to 25.

    \item Utilization of statistical modes to establish the ultimate ranking of each risk factor, providing valuable insights into the perceived severity of MSD risks.

    \item Contribution to the field of MSD research by providing a robust framework for risk factor classification and ranking, which can aid in informed decision-making for MSD prevention and management.

    \item Practical implications for individuals, organizations, and policymakers in mitigating the impact of MSD and enhancing workplace ergonomics.

\end{itemize}

To categorize risk variables for MSD, this research introduces a novel method that combines distance measurements with sophisticated NLP pre-trained transformer models. It is significant since it is the first attempt to use NLP techniques to analyze risk variables for MSDs. Rather than the traditional three-category classification encompassing personal, physical, and psychological factors (\citealp{haukka_mental_2011}; \citealp{kumar_information_2018}; \citealp{bispo_risk_2022}; \citealp{tang_prevalence_2022}), this study ventures to classify these factors into five distinct categories: personal, biomechanical, workplace, psychological, and organizational. Scientific data supports this change by indicating that physical risk factors may be further subdivided into biomechanical and occupational factors, providing a more nuanced understanding. Similar segments exist for psychological risk factors: organizational and psychological. In addition, this study is the first to assess the relative importance of MSD risk variables using a descriptive statistical analysis of workers' perspectives, offering crucial new information to people, society, and organizations.

The article is structured as follows: in the ``\hyperref[sec:lr]{Literature review}" section, we meticulously examined previously identified MSD risk factors in the literature. From this review, we distilled 25 unique MSD risk factors, which we further organized into 5 broad categories. These categories serve as the target features for the NLP models employed in our study. The ``\hyperref[sec:methods]{Methodology}" section outlines the methodology employed for categorizing these risk factors through the application of NLP techniques, delves into the design process of our MSD risk factors ranking survey, and elucidates our approach to ranking them. Furthermore, the ``\hyperref[sec:result]{Results and discussions}" section critically analyzes the classification performance and their statistical validation tests of the NLP models, explores the significance of identifying the most critical risk factors, and discusses the managerial implications of our AI-driven approach. The ``\hyperref[sec:conclusion]{Conclusions and future research directions}" section encapsulates our novel findings, contributions, and limitations and sets the stage for future research directions.

\section{Literature review}
\label{sec:lr}

\begin{table*}[!ht]
\centering
\caption{Identified 25 MSD risk factors from comprehensive literature review}
\label{tab:msd-risks}
\resizebox{\linewidth}{!}{%
\begin{tabular}{|>{\raggedright\arraybackslash}p{0.15\linewidth}|l|m{0.5\linewidth}|} 
\hline
\textbf{Category}                                         & \textbf{MSD risk factor}& \textbf{References}                                                                                                                                                                                                                              \\ 
\hline
\multirow{5}{0.075\linewidth}{\hspace{0pt}Personal}       & Age                                   & \cite{barlas_individual_2018,sekkay_risk_2018,yang_stock_2021,bispo_risk_2022,caporale_assessing_2022,kashif_work-related_2022,kim_cross-sectional_2022,shohel_parvez_assessment_2022,tang_prevalence_2022}  \\ 
\cline{2-3}
                                                          & Gender                                & \cite{sekkay_risk_2018,yang_stock_2021,bispo_risk_2022,caporale_assessing_2022,tang_prevalence_2022,shohel_parvez_assessment_2022}                                                                                 \\ 
\cline{2-3}
                                                          & Anthropometry                         & \cite{yang_stock_2021,shohel_parvez_assessment_2022,tang_prevalence_2022}                                                                                                                                                \\ 
\cline{2-3}
                                                          & Lifestyle                             & \cite{sekkay_risk_2018,yang_stock_2021,kashif_work-related_2022,srivastava_musculoskeletal_2022,tang_prevalence_2022,laflamme_is_2004,phonrat_risk_1997,ueno_association_1999}                                  \\ 
\cline{2-3}
                                                          & Work Experience                       & \cite{barlas_individual_2018,kashif_work-related_2022,mahmud_employee_2023}                                                                                                                                               \\ 
\hline
\multirow{5}{0.075\linewidth}{\hspace{0pt}Workplace}      & Layout                                & \cite{barlas_individual_2018,kim_cross-sectional_2022,shohel_parvez_assessment_2022,ruzairi_systematic_2022,bispo_risk_2022}                                                                                         \\ 
\cline{2-3}
                                                          & Pace of Work                          & \cite{kaya_aytutuldu_musculoskeletal_2022,barlas_individual_2018,kim_cross-sectional_2022,joshi_study_2022}                                                                                                            \\ 
\cline{2-3}
                                                          & Noise                                 & \cite{caporale_assessing_2022,kim_cross-sectional_2022,stansfeld_noise_2003}                                                                                                                                              \\ 
\cline{2-3}
                                                          & Inappropriate Lighting                & \cite{aksut_determining_2023,caporale_assessing_2022,srivastava_musculoskeletal_2022}                                                                                                                                     \\ 
\cline{2-3}
                                                          & Environmental Condition               & \cite{aksut_determining_2023,caporale_assessing_2022,srivastava_musculoskeletal_2022,joshi_study_2022,barlas_individual_2018,shohel_parvez_assessment_2022}                                                        \\ 
\hline
\multirow{5}{0.075\linewidth}{\hspace{0pt}Psycholosocial} & Job dissatisfaction                   & \cite{bispo_risk_2022}; \cite{yang_stock_2021}                                                                                                                                                             \\ 
\cline{2-3}
                                                          & Social support                        & \cite{heijden_prevalence_2019}; \cite{ruzairi_systematic_2022}                                                                                                                                             \\ 
\cline{2-3}
                                                          & Mental and occupational stress        & \cite{oubibi_perceived_2022}; \cite{sutarto_exploring_2022}                                                                                                                                                \\ 
\cline{2-3}
                                                          & Job insecurity                        & \cite{mahmud_employee_2023}; \cite{bispo_risk_2022}                                                                                                                                                        \\ 
\cline{2-3}
                                                          & Effort-reward imbalance               & \cite{jiskani_mine_2020}; \cite{newman_experiences_2022}                                                                                                                                                   \\ 
\hline
\multirow{5}{0.075\linewidth}{\hspace{0pt}Organizational} & Insufficient breaks                   & \cite{shohel_parvez_assessment_2022}; \cite{joshi_study_2022}; \cite{sekkay_risk_2018}                                                                                                 \\ 
\cline{2-3}
                                                          & Poor job design                       & \cite{bazazan_association_2019}; \cite{jiskani_mine_2020}; \cite{bispo_risk_2022}                                                                                                       \\ 
\cline{2-3}
                                                          & High job demand                       & \cite{l_kalleberg_probing_2017}; \cite{keyaerts_association_2022}                                                                                                                                         \\ 
\cline{2-3}
                                                          & Management style                      & \cite{tang_prevalence_2022}; \cite{jiskani_mine_2020}                                                                                                                                                      \\ 
\cline{2-3}
                                                          & Poor employee facilities              & \cite{roquelaure_personal_2020}; \cite{yang_stock_2021}                                                                                                                                                    \\ 
\hline
\multirow{5}{0.075\linewidth}{\hspace{0pt}Biomechanical}  & Working Posture                       & \cite{smith_management_2022,roquelaure_personal_2020,caporale_assessing_2022,bispo_risk_2022,joshi_application_2023}                                                                                                  \\ 
\cline{2-3}
                                                          & Vibration                             & \cite{sekkay_risk_2018,kim_cross-sectional_2022,shohel_parvez_assessment_2022,ruzairi_systematic_2022}                                                                                                                 \\ 
\cline{2-3}
                                                          & Repetitive Motion                     & \cite{kim_cross-sectional_2022,tang_prevalence_2022}                                                                                                                                           \\ 
\cline{2-3}
                                                          & Force                                 & \cite{ruzairi_systematic_2022,sekkay_risk_2018,kim_cross-sectional_2022,shohel_parvez_assessment_2022}                                                                                                                 \\ 
\cline{2-3}
                                                          & Deviation from Neutral Body Alignment & \cite{joshi_application_2023,bispo_risk_2022}                                                                                                                                                                               \\
\hline
\end{tabular}
}
\end{table*}

The purpose of this review of the research is to clarify how MSD risk variables are arranged hierarchically. The thorough evaluation procedure included classifying the risk factors for MSDs using various academic sources, including articles indexed by Scopus, Web of Science, and Google Scholar. To organize our review, we categorized the literature into two main classes. During the first search, we carefully examined scholarly articles in Web of Science and Scopus. On the other hand, the second search included reports, unpublished (gray) material, and relevant publications that well-known organizations did not index. Our search strategy employed a range of keywords, including `MSD risk factors classification,' `MSD personal risk factors classification,' `MSD individual risk factors classification,' `MSD biomechanical risk factors classification,' `MSD ergonomic risk factors classification,' `MSD organizational risk factors classification,' `MSD psychological risk factors classification,' `MSD psychosocial risk factors classification,' and `workplace risk factors classification of MSD,' among others. We limited the scope of our first screening to articles written in fluent English released in the previous ten years (2012–2024). The final selection included publications that discussed various risk factors for MSDs as well as their relationship to MSD issues. Our first keyword search produced an enormous corpus of more than 1180 articles. These were then narrowed down by removing duplicates and unrelated content, leaving us with a carefully chosen corpus of 157 legitimate articles. To be as accurate as possible, we carefully considered each abstract before choosing one. Unfortunately, several papers that only mentioned MSD risk factors without providing evidence of their strong association with MSD were disqualified at the last screening stage. Ultimately, we distilled our selection to 21 studies that comprehensively elucidated the risk factors associated with MSD problems, as shown in Table \ref{tab:msd-risks}.

Personal risk factors associated with MSD encompass a spectrum of individual physical attributes. From prior literature, several noteworthy risk factors in this category are evident, including age, gender, anthropometry, lifestyle, and work experience. The realm of physical risk factors associated with MSD problems encompasses a spectrum of biomechanical elements intertwined with the workplace environment \citep{sekkay_risk_2018,kim_cross-sectional_2022, tang_prevalence_2022}. Drawing from pertinent literature, a selection of exemplary risk factors includes working posture, repetitive motion, force application, deviations from neutral body alignment, exposure to vibrations, workplace layout, pace of work, suboptimal lighting, environmental conditions, and noise \citep{sekkay_risk_2018, bispo_risk_2022, ruzairi_systematic_2022,joshi_study_2022}. Psychological risk factors are intricately linked to the psychosocial aspects of workers, organizations' managerial practices, and employee support policies. Drawing from previous literature, noteworthy examples of these psychological risk factors encompass job dissatisfaction, social support, mental and occupational stress, job insecurity, effort-reward imbalance, insufficient breaks, poor job design, high job demand, management style, and the facilities provided to employees \citep{sekkay_risk_2018,yang_stock_2021,bispo_risk_2022,sutarto_exploring_2022,ruzairi_systematic_2022,bazazan_association_2019,shohel_parvez_assessment_2022,keyaerts_association_2022}.

Pereira et al. developed a new classification scheme combining occupational medicine criteria with principles from clinical pathology for work-related MSDs (WRMSDs) \citep{pereira_musculoskeletal_2021}. This all-inclusive method, approved by specialists, provides a useful framework for identifying and treating WRMSDs and encourages early detection and prevention to lessen chronicity. Graded into four categories according to inflammatory processes and modalities of injury start, the classification improves communication between medical experts from various specializations, leading to better treatment outcomes and a decrease in the incidence of industrial injuries. Sasikumar \& Binoosh aimed to create a prediction model for computer workers' MSD risk. An analysis was conducted on postural, physiological, and work-related aspects using a modified Nordic questionnaire and quick upper limb assessment \citep{sasikumar_model_2020}. The accuracy of machine learning techniques, such as Random Forest and Naïve Bayes Classifier, in predicting the risk of MSDs was high (81.25\%). The development of MSD was significantly influenced by posture, physiology, and work-related variables; this emphasizes the significance of taking preventative steps to lower occupational health risks for computer workers. Using structural equation modeling, Talapatra et al. examined the effects of several risk variables on the emergence of MSDs \citep{talapatra_assessing_2023}. They found that biomechanical, occupational, psychological, personal, and organizational risk factors significantly impact MSD issues. The results emphasized how critical it is to address these factors separately or in combination to reduce the incidence of MSDs. Twenty-five risk factors for MSDs were carefully identified and categorized into five divisions by Talapatra et al.: personal, biomechanical, occupational, psychological, and organizational \citep{talapatra_musculoskeletal_2022}. The study ranked these parameters using the Fuzzy Analytical Hierarchy Process (FAHP), and the top five important risk factors were vibration, anthropometry, working position, repetitive motion, and layout. Contreras-Valenzuela \& Martínez-Ibanez found that workers' knees were the most affected body parts, with 47 cases. The resulting work-related MSDs included tendinitis, arthralgia, chondromalacia, and gonarthrosis \citep{contreras-valenzuela_hierarchical_2024}. The musculoskeletal stress factor with the most significant impact on the body’s health was fatiguing work, which involved repeated jumps, prolonged squatting, or kneeling, present in cluster 1 and cluster 3. When the back was mildly flexed forward with one leg used more often in supporting the body, the repeated work position had the highest frequency of 63 and 56 answers. Li et al. investigated a brand-new deep learning-based rapid upper limb assessment (RULA) end-to-end implementation \citep{li_novel_2020}. It predicted the RULA action level, a subset of the RULA grand score, using input from regular RGB photos. Lifting postures measured in the lab and posture data from the Human 3.6 dataset were used for training and assessment. The algorithm's performance in identifying the RULA action level was 93\% accurate and 29 frames per second efficient. Additionally, the outcomes showed that using data augmentation significantly improved the model's resilience. The study emphasizes how the suggested real-time on-site risk assessment approach might help reduce MSD associated with the workplace.

A range of studies have explored using NLP in categorizing risk factors. Khalifa adapted existing NLP tools to identify cardiovascular risk factors in clinical notes, achieving an F1-measure of 87.5\% \citep{khalifa_adapting_2015}. Madeira developed a methodology using NLP for predicting human factors in aviation incidents, achieving a Micro F1 score of 90\% \citep{madeira_machine_2021}. Chen developed a hybrid pipeline system for identifying heart disease risk factors in clinical texts, achieving an F1-score of 92.68\% \citep{chen_automatic_2015}. Xue proposed a method for determining risk factors using multilayer neural networks as a classifier \citep{xue_searching_1996}. These studies collectively demonstrate the potential of NLP in risk factor classification with high levels of accuracy and efficiency. Arora developed a machine learning model to predict MSDs in garment industry workers, achieving an accuracy of 91.3\% \citep{arora_machine_2021}. This approach could be extended to other industries and populations. Halsey and Brooks emphasized the importance of early intervention and provided guidelines for diagnosing and managing MSDs \citep{halsey_musculoskeletal_2004,brooks_template_1996}. However, these studies did not specifically focus on NLP-based risk factor classification.

The literature gap is evident in the complete absence of NLP utilization within ergonomics and occupational health research. Despite the well-documented prevalence and impact of MSDs in various industries, there is a notable absence of studies that have leveraged NLP techniques to analyze and understand MSD risk factors. This represents a significant oversight, as NLP holds immense potential for transforming how researchers extract insights from textual data, including scientific literature, medical records, and expert opinions. By not tapping into the capabilities of NLP, the field misses out on opportunities to enhance the depth and breadth of its research findings, ultimately hindering progress in the prevention and management of MSDs. Addressing this gap requires a concerted effort to integrate NLP methodologies into ergonomics and occupational health research, thereby unlocking new avenues for advancing our understanding of MSD risk factors and improving workplace health and safety practices.

\section{Methodology}
\label{sec:methods}
\begin{figure*}[!ht]
    \centering
    \includegraphics[width=1\linewidth]{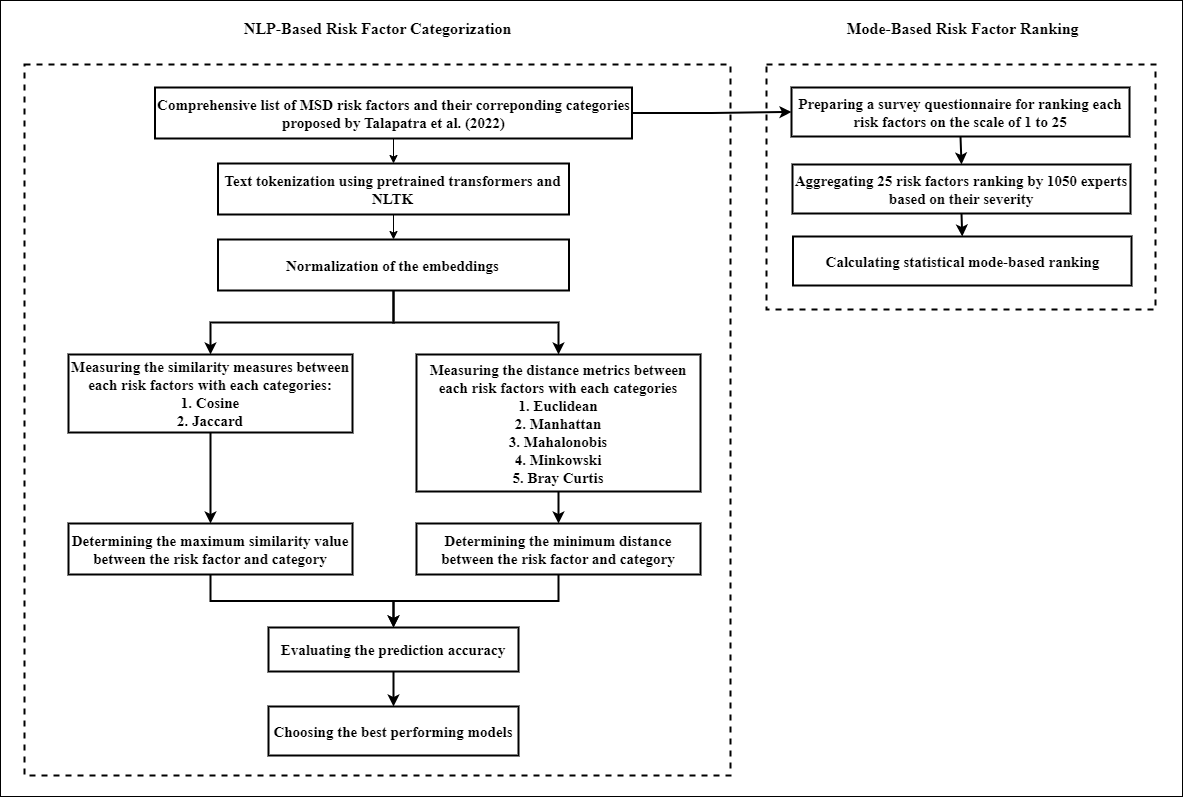}
    \caption{Methodological framework illustrating NLP-based risk factor classification (left) and statistical mode-based risk factor ranking (right).}
    \label{fig4}
\end{figure*}

\subsection{Risk factor classification using NLP}
\label{sec:risk_factor_cat}
This subsection details the methodology employed for categorizing risk factors extracted from the literature review using NLP techniques, as shown in Figure \ref{fig4}. The objective was to assign predefined labels to these risk factors based on their contextual similarity to establish a comprehensive classification framework. This framework is an essential component of our research, as it forms the foundation for the subsequent ranking and analysis of risk factors by severity.

\subsubsection{Data collection}
The dataset for this classification task was curated through an extensive literature review in occupational health and safety. A total of 25 risk factor phrases were extracted from peer-reviewed research articles, industry reports, and authoritative sources (\citep{talapatra_musculoskeletal_2022}). These risk factors represent various aspects of occupational hazards, ranging from ergonomic concerns to psychosocial factors. Each of the 25 risk factors was manually assigned one of five predefined labels to facilitate the NLP-based classification. These labels correspond to the overarching categories: ``personal," ``workplace," ``psychosocial," ``organizational," and ``biomechanical." Domain experts conducted this labeling process to ensure accuracy and consistency.

\subsubsection{NLP model selection}
We adopted four different approaches for NLP model selection: BERT with cosine similarity, NLTK with Jaccard similarity, sentence transformer with cosine similarity, and sentence transformer with distance metrics. In this research, the selection of specific models, namely NLTK, BERT, and sentence transformer, is grounded in each model's unique strengths and capabilities for their respective tasks. NLTK, a comprehensive library for NLP in Python, is chosen for its versatility and established tools, making it suitable for preliminary text processing and linguistic analysis. BERT, renowned for its deep contextualized representations, is employed for its superior performance in understanding the intricate context and semantics of MSD risk factors, ensuring a nuanced classification. The sentence transformer, designed to generate meaningful sentence embeddings, complements BERT by providing a simplified and efficient representation for mode-based ranking, enhancing the quantitative analysis of risk factors. This thoughtful combination of NLTK, BERT, and sentence transformer reflects a holistic approach, leveraging the specific strengths of each model to address different aspects of the research tasks effectively.

\begin{figure*}
    \centering
    \includegraphics[width=0.95\linewidth, height=1.25\linewidth]{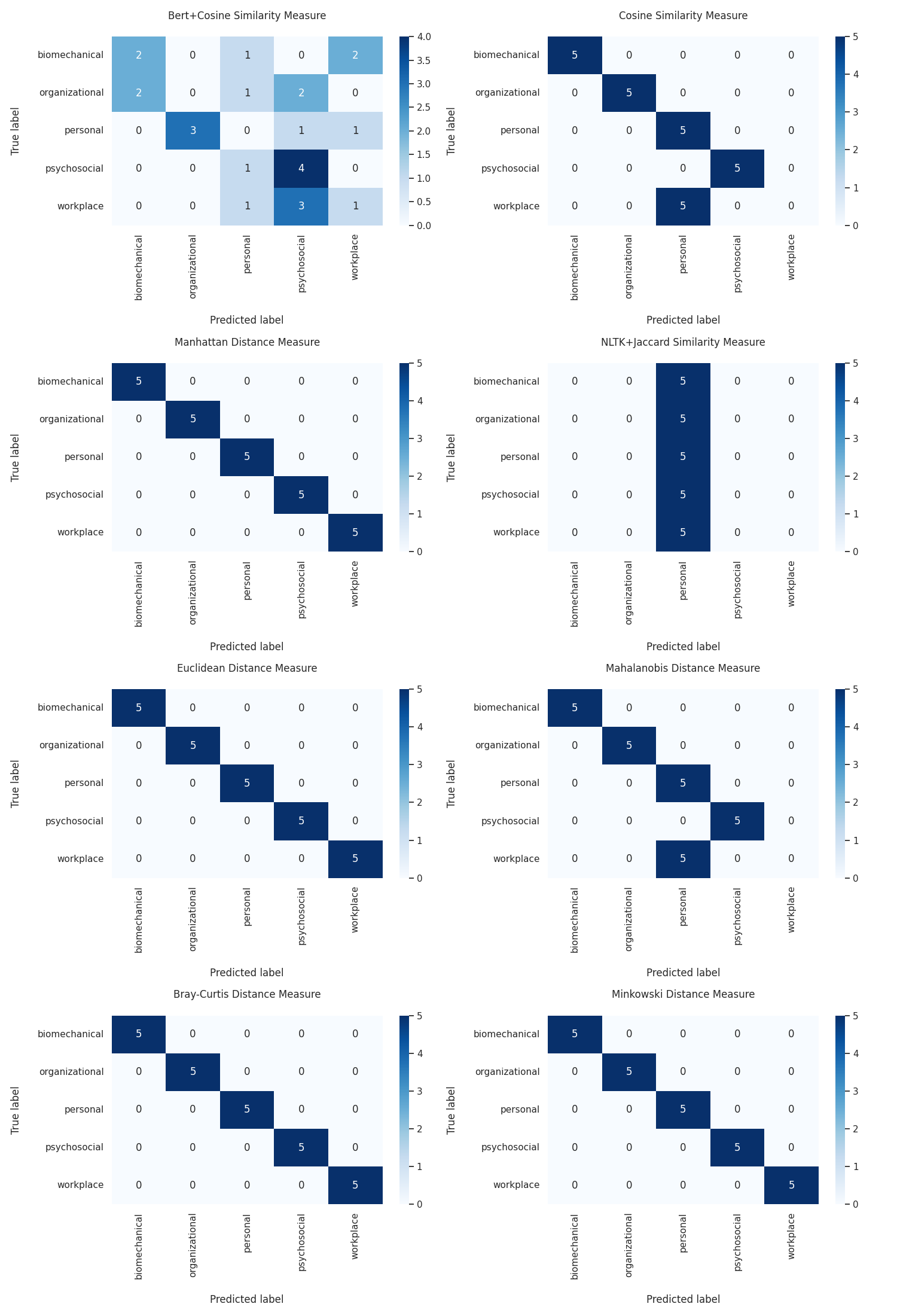}
    \caption{Confusion matrices for 8 NLP models employed for risk factor classification, including BERT + cosine similarity measure, cosine similarity measure, Manhattan distance measure, NLTK + Jaccard similarity measure, Euclidean distance measure, Mahalanobis distance measure, Bray-Curtis distance measure, and Minkowski distance measure, listed from top left to bottom right.}
    \label{fig1}
\end{figure*}

\begin{figure*}[!ht]
    \centering
    \includegraphics[width=1\linewidth, height=1.2\linewidth]{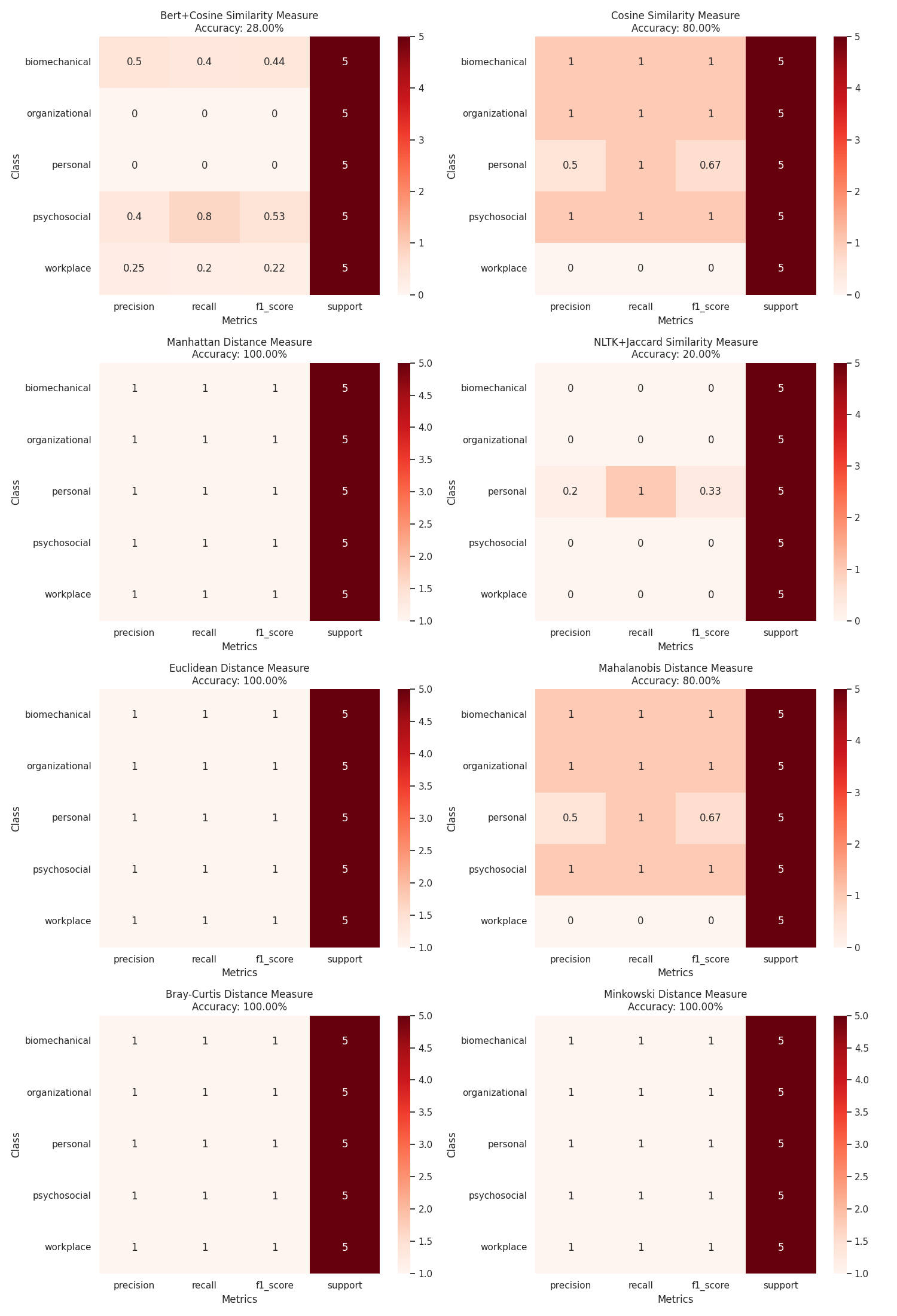}
    \caption{Classification reports for 8 NLP models employed for risk factor classification, including BERT + cosine similarity measure, cosine similarity measure, Manhattan distance measure, NLTK + Jaccard similarity measure, Euclidean distance measure, Mahalanobis distance measure, Bray-Curtis distance measure, and Minkowski distance measure, listed from top left to bottom right.}
    \label{fig2}
\end{figure*}

\textbf{BERT with Cosine Similarity:}
In the NLP domain, we harnessed the power of the Bidirectional Encoder Representations from Transformers (BERT) model, a pre-trained neural network known for its remarkable capabilities in understanding textual data. Specifically, we employed the `bert-base-uncased' configuration and the corresponding tokenizer to preprocess text inputs. The `AutoModel' and `AutoTokenizer' modules from the transformers library facilitated the efficient loading and configuration of the BERT model.

Cosine similarity, a widely recognized metric for quantifying the semantic similarity between textual elements, was chosen to compute the similarity between each risk factor and the predefined categories. It measures the cosine of the angle between two vectors in a multi-dimensional space. It quantifies the similarity between vectors based on the cosine of the angle formed by the vectors. Cosine similarity produces values between -1 and 1, with larger values indicating greater similarity. The cosine similarity can be computed using the dot product and vector norms:
\begin{equation}
  \text{Cosine Similarity}(\mathbf{A}, \mathbf{B}) = \frac{\sum_{i=1}^{n} A_i \cdot B_i}{\sqrt{\sum_{i=1}^{n} A_i^2} \cdot \sqrt{\sum_{i=1}^{n} B_i^2}}  
\label{eq1}
\end{equation}
Where $A$ and $B$ represent vectors representing risk factors or label embeddings, and $n$ variable represents the dimensionality of the vectors $A$ and $B$, which is the number of elements or features in each vector.

Our model predicted the appropriate label for each risk factor by identifying the category with the highest similarity score. The `argmax' function was used to extract the index with the highest similarity score, thus determining the predicted label for each risk factor.

\textbf{NLTK with Jaccard Similarity:}
In this step, Python's Natural Language Toolkit (NLTK) library is used to tokenize the risk factors and labels. Tokenization breaks down sentences or text into individual words or tokens, making processing and analyzing text data easier. For each risk factor and label in your dataset, the `word\_tokenize` function is applied to split them into lists of words or tokens. The Jaccard similarity measures how similar two sets are. In our case, it is used to measure the similarity between the tokens of risk factors and labels. The Jaccard similarity between two sets A and B is defined as the size of their intersection divided by the size of their union:
\begin{equation}
    J(A, B) = \frac{|A \cap B|}{|A \cup B|}
\end{equation}

After calculating Jaccard similarities for each pair of risk factors and label, the predicted label is determined by selecting the label with the highest Jaccard similarity. This is done for each risk factor.

\textbf{Sentence Transformer with Cosine Similarity:}
We extended our NLP capabilities by employing the `paraphrase-MiniLM-L6-v2' pre-trained SentenceTransformer model. This model facilitated the generation of embeddings for both labels and risk factors.

We computed the cosine similarity between the normalized embeddings of risk factors and labels to measure the similarity between risk factors and labels. This process involved calculating the dot product between the normalized risk factor embeddings and the normalized label embeddings. The resulting similarities were transformed into a NumPy array for further analysis.

\textbf{Sentence Transformer with Distance Metrics:}
The above-mentioned sentence transformer model was loaded, and subsequently, $L2$ normalization was applied to both label and risk factor embeddings to ensure consistency in similarity calculations. Normalization scales the embeddings to have unit length, ensuring their magnitudes do not affect subsequent distance calculations. The L2 normalization of X (denoted as X') is calculated as:

\begin{equation}
    {X'} = \left[\frac{x_1}{\|X\|}, \frac{x_2}{\|X\|}, \frac{x_3}{\|X\|}, \ldots, \frac{x_i}{\|X\|}\right]
\end{equation}

Where $\|{X}\| = \sqrt{x_1^2 + x_2^2 + x_3^2 + \ldots + x_i^2}$; $X'$ denotes the normalized vector after L2 normalization, and $x_i$ denotes an individual element in the original vector.

We incorporated five distinct distance metrics to compute the distances between embeddings: Euclidean distance, Manhattan distance, Mahalanobis distance, Bray-Curtis distance, and Minkowski distance with p=3. These distances were computed using appropriate functions from the `scipy.spatial.distance' library. The label with the minimum distance was predicted for each risk factor. Based on the computed distances, this predicted the most similar label for each risk factor.

Euclidean distance measures the straight-line distance between two points in a multi-dimensional space. It quantifies dissimilarity, with smaller distances indicating greater similarity.

\begin{equation}
    \text{Euclidean Distance}(\mathbf{A}, \mathbf{B}) = \sqrt{\sum_{i=1}^{n} (A_i - B_i)^2}
\end{equation}

The Euclidean distance can be computed as the square root of the sum of squared differences between corresponding elements of vectors \({A}\) and \({B}\).

Manhattan distance, also known as the L1 distance, calculates the sum of absolute differences between corresponding elements of two vectors. It measures dissimilarity by summing the absolute differences in each dimension.

\begin{equation}
    \text{Manhattan Distance}(\mathbf{A}, \mathbf{B}) = \sum_{i=1}^{n} |A_i - B_i|
\end{equation}

The Manhattan distance can be computed as the sum of absolute differences between corresponding elements of vectors \({A}\) and \({B}\).

Mahalanobis Distance is a measure of the distance between a point or vector $x$ and a distribution or set of data points with mean \(\boldsymbol{\mu}\) and covariance matrix \(\Sigma\). It takes into account the correlations between variables, making it useful when dealing with multivariate data. The formula for Mahalanobis distance is as follows:

\begin{equation}
    \text{Mahalonobis Distance} = \sqrt{(\mathbf{x} - \boldsymbol{\mu})^\top \Sigma^{-1} (\mathbf{x} - \boldsymbol{\mu})}
\end{equation}

Where \(\mathbf{x}\) represents the data point or vector you want to measure the distance from the distribution. \(\boldsymbol{\mu}\) is the mean vector of the distribution. \(\Sigma\) is the covariance matrix of the distribution. \(\Sigma^{-1}\) is the inverse of the covariance matrix. Mahalanobis distance was used to measure the dissimilarity or proximity between risk factors, taking into account their multivariate nature and correlations.

The Minkowski distance is a generalization of several distance metrics, including Euclidean and Manhattan distances. It is defined as:

\begin{equation}
    \text{Minkowski Distance}(\mathbf{A}, \mathbf{B}) = \left(\sum_{i=1}^{n} |A_i - B_i|^p\right)^{1/p}
\end{equation}

Where $p$ is a parameter that determines the type of distance metric. When \(p = 2\), it becomes the Euclidean distance; when \(p = 1\), it becomes the Manhattan distance. It allows you to adjust the sensitivity of the distance metric to different dimensions.

The Bray-Curtis dissimilarity is a metric for comparing the compositional dissimilarity between two sets. It is often used in ecology to measure dissimilarity between ecological communities. It is defined as:

\begin{equation}
    \text{Bray Curtis Distance}(\mathbf{A}, \mathbf{B}) = \frac{\sum |A_i - B_i|}{\sum |A_i + B_i|}
\end{equation}

Where \(A_i\) and \(B_i\) represent the abundance or composition of the \(i\)-th component in two sets \(A\) and \({B}\).

The computed distances between normalized risk factor embeddings and normalized label embeddings were gathered for each distance metric. The Euclidean, Mahalanobis, Bray-Curtis, and Minkowski distances were transposed for analytical purposes. This transformation was done to structure the data for further in-depth analysis and interpretation.

\subsection{Ranking of risk factors}
\label{sec:ranking}
\subsubsection{Survey design for ranking}
In order to establish the severity ranking of the identified MSD risk factors, a carefully structured survey was designed, as shown in Figure \ref{fig4}. The survey aimed to gather expert respondents' insights regarding each risk factor's perceived severity. The survey utilized a numerical scale ranging from 1 to 25, with 1 signifying the highest perceived severity and 25 representing the lowest. This scale allowed participants to express their judgment regarding each risk factor's seriousness numerically. 

The survey presented the list of 25 identified risk factors to the participants, who were then asked to assign a severity rating to each one. The design ensured that participants could clearly understand and distinguish between the risk factors, enabling them to provide informed rankings based on their perceptions. The datasets are publicly available in ``\href{https://data.mendeley.com/datasets/kr33mvtw63/1}{Survey Data on Ranking of Musculoskeletal Disorder Risk Factors}" \citep{jahin_msd_2024} to ensure reproducibility.

A total of 1050 participants were engaged in this survey, forming a diverse and representative sample. The survey participants encompassed individuals from various demographic backgrounds, including different age groups, genders, educational levels, and professional experiences. This diverse pool of respondents was essential to capture a comprehensive perspective on the severity ranking of MSD risk factors.\\
\textbf{Age Groups:} The participant pool covered a wide range of age groups, including individuals in their twenties, thirties, forties, and beyond. This diversity allowed for a multi-generational perspective on MSD risk factor severity.\\
\textbf{Genders:} The survey included respondents of different genders, fostering a gender-inclusive approach to understanding and ranking MSD risk factors. This ensured a nuanced evaluation of how gender might influence perceptions of risk severity.\\
\textbf{Educational Levels:} Participants had varying educational backgrounds, ranging from individuals with undergraduate degrees to those holding advanced degrees such as master's and doctoral qualifications. This diversity in education levels contributed to a well-rounded evaluation of risk factors.\\
\textbf{Professional Experiences:} The participant pool consisted of professionals with diverse career experiences, including but not limited to:
\begin{itemize}
    \item Academic Professionals: Professors, lecturers, and researchers from esteemed universities and research institutions.
    
    \item Industry Experts: Individuals with extensive experience in various industries, providing practical insights into occupational contexts.
    
    \item Healthcare Practitioners: Doctors, physiotherapists, and health professionals specializing in musculoskeletal health.
    
    \item Corporate Professionals: Employees from different sectors, bringing insights from corporate environments.
\end{itemize}
\textbf{Geographical Representation:} Participants were geographically dispersed, representing various regions and countries. This global perspective ensured the consideration of cultural and occupational differences in evaluating MSD risk factors.

This strategic selection of participants with diverse backgrounds enhances the external validity of the study, making the findings applicable to a broad range of contexts and populations. Including participants from different walks of life strengthens the robustness and comprehensiveness of the research outcomes.

\begin{figure*}[!ht]
    \centering
    \includegraphics[width=1\linewidth, height=1.1\linewidth]{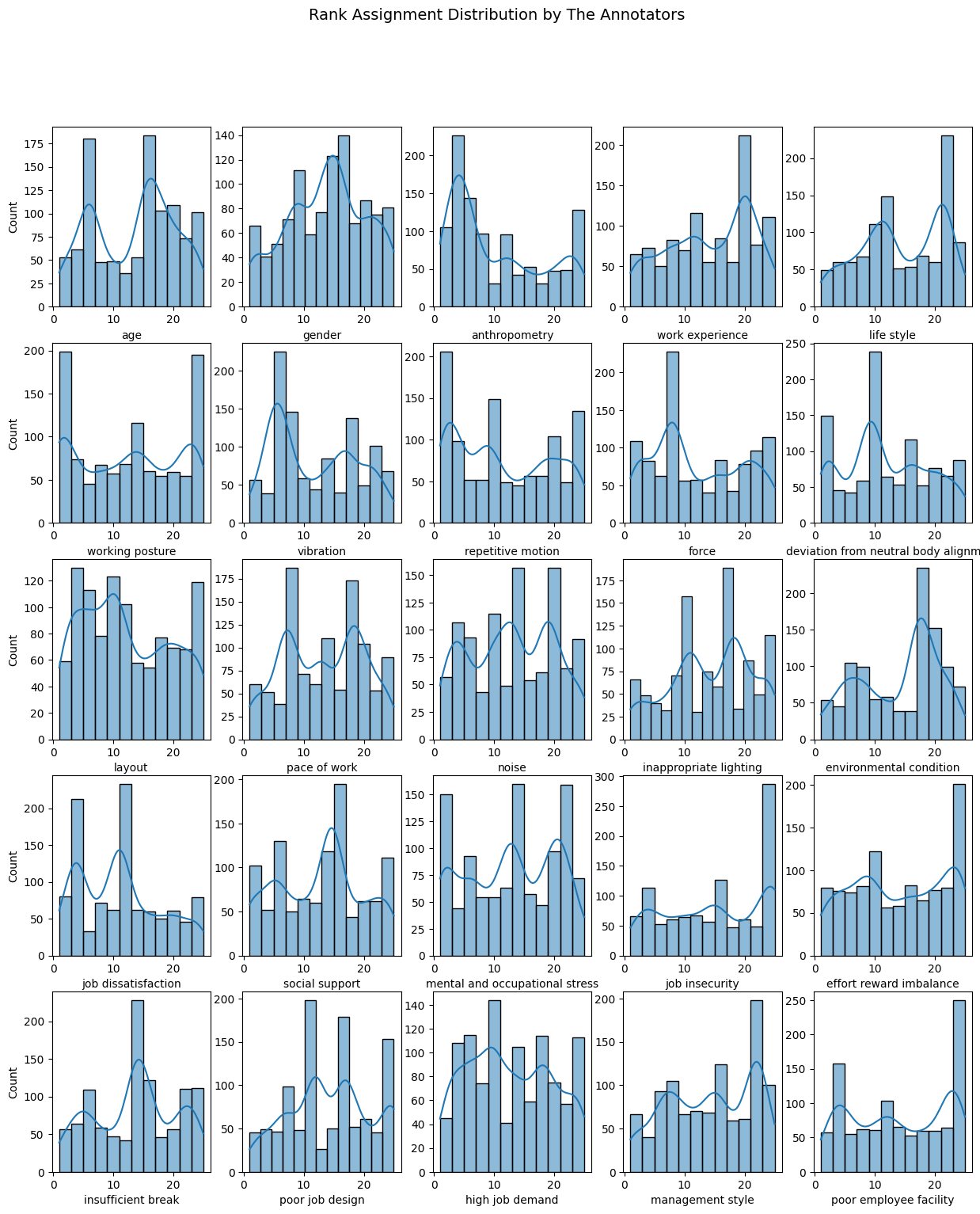}
    \caption{Distribution of severity rankings of the MSD risk factors on the numerical scale ranging from 1 to 25.}
    \label{fig3}
\end{figure*}

\subsubsection{Rank calculation}

The survey responses were meticulously analyzed to determine the ultimate ranking of each MSD risk factor by severity. The calculation process involved computing the mode for each risk factor. The mode, representing the most frequently occurring ranking for a particular risk factor across all responses, was chosen as the definitive ranking score.

The rationale behind selecting the mode was to ensure that the ranking reflected the collective perception of the survey participants. By identifying the ranking most commonly assigned to each risk factor, we aimed to understand which risk factors were consistently regarded as more severe by the respondents.

The mode-based ranking approach facilitated the identification of the most critical MSD risk factors, allowing for targeted intervention and prevention strategies.

\begin{table*}[!ht]
\centering
\caption{Mode-based rankings of MSD risk factors based on survey}
\begin{tabular}{lc} 
\toprule
\textbf{MSD Risk Factors} & \textbf{Rank} \\ 
\hline
Age & 6 \\
Anthropometry & 4 \\
Deviation from neutral body alignment & 9 \\
Effort reward imbalance & 24 \\
Environmental condition & 17 \\
Force & 8 \\
Gender & 16 \\
High job demand & 10 \\
Inappropriate lighting & 18 \\
Insufficient break & 14 \\
Job dissatisfaction & 11 \\
Job insecurity & 25 \\
Layout & 3 \\
Lifestyle & 21 \\
Management style & 22 \\
Mental and occupational stress & 13 \\
Noise & 19 \\
Pace of work & 7 \\
Poor employee facility & 23 \\
Poor job design & 12 \\
Repetitive motion & 2 \\
Social support & 15 \\
Vibration & 5 \\
Work experience & 20 \\
Working posture & 1 \\
\bottomrule
\end{tabular}
\label{tab1}
\end{table*}

\begin{table*}[!ht]
\centering
\caption{Descriptive statistics of the surveyed MSD risk factor rankings}
\begin{tabular}{lccccccc} 
\toprule
\multicolumn{1}{c}{\textbf{MSD Risk Factors}} & \textbf{Mean} & \textbf{Standard Deviation} & \textbf{Minimum} & \textbf{25\%} & \textbf{50\%} & \textbf{75\%} & \textbf{Maximum} \\ 
\hline
Age & 13.35 & 6.81 & 1 & 6 & 15 & 19 & 25 \\
Anthropometry & 10.26 & 7.51 & 1 & 4 & 7 & 16 & 25 \\
Deviation From Neutral Body
  Alignment & 11.96 & 6.91 & 1 & 7 & 10 & 18 & 25 \\
Effort Reward Imbalance & 13.74 & 7.64 & 1 & 7 & 14 & 21 & 25 \\
Environmental Condition & 13.96 & 6.64 & 1 & 8 & 17 & 19 & 25 \\
Force & 12.12 & 7.32 & 1 & 7 & 10 & 19 & 25 \\
Gender & 13.81 & 6.52 & 1 & 9 & 14 & 19 & 25 \\
High Job Demand & 12.62 & 6.93 & 1 & 6 & 12 & 18 & 25 \\
Inappropriate Lighting & 14.27 & 6.71 & 1 & 10 & 15 & 19 & 25 \\
Insufficient Break & 13.57 & 6.79 & 1 & 8 & 14 & 19 & 25 \\
Job Dissatisfaction & 11.00 & 6.72 & 1 & 4 & 11 & 16 & 25 \\
Job Insecurity & 14.66 & 8.03 & 1 & 8 & 15 & 23 & 25 \\
Layout & 12.14 & 7.19 & 1 & 6 & 11 & 18 & 25 \\
Life Style & 14.18 & 6.79 & 1 & 9 & 14 & 21 & 25 \\
Management Style & 14.09 & 7.08 & 1 & 8 & 15 & 22 & 25 \\
Mental And Occupational Stress & 12.69 & 7.33 & 1 & 6 & 13 & 19 & 25 \\
Noise & 12.85 & 6.88 & 1 & 7 & 13 & 19 & 25 \\
Pace Of Work & 13.22 & 6.61 & 1 & 7 & 13 & 18 & 25 \\
Poor Employee Facility & 13.79 & 7.76 & 1 & 6 & 13 & 22 & 25 \\
Poor Job Design & 14.36 & 6.78 & 1 & 9 & 14 & 19 & 25 \\
Repetitive Motion & 11.46 & 7.75 & 1 & 4 & 9 & 19 & 25 \\
Social Support & 12.62 & 7.00 & 1 & 6 & 14 & 17 & 25 \\
Vibration & 11.76 & 6.84 & 1 & 5 & 10 & 17 & 25 \\
Work Experience & 14.12 & 7.00 & 1 & 8 & 15 & 20 & 25 \\
Working Posture & 12.41 & 8.22 & 1 & 4 & 13 & 20 & 25 \\
\bottomrule
\end{tabular}
\end{table*}

\section{Results and discussions}
\label{sec:result}
In this section, we delve into the results obtained from our study on the classification and ranking of MSD risk factors, providing an in-depth analysis of the performance of different models and the implications of the ranking results.

\subsection{Analysis of classification performance}
In Figure \ref{fig1}, we present the confusion matrices for a comprehensive evaluation of eight distinct approaches employed in the classification of MSD risk factors. Each matrix showcases the performance of a specific approach in categorizing risk factors into their relevant classes: personal, biomechanical, workplace, psychological, and organizational. The diagonal elements depict accurately classified instances, while off-diagonal elements indicate instances of misclassification. These matrices provide valuable insights into the effectiveness of each approach in identifying MSD risk factors.

This Figure \ref{fig2} and Table \ref{tab3} presents the comprehensive classification reports that offer a detailed assessment of the models' performance, including precision, recall, F1-score, and support metrics for both positive and negative classes. These evaluation scores were computed by considering the classification by \citep{talapatra_musculoskeletal_2022}. The classification reports serve as a critical reference for understanding the strengths and weaknesses of each model in the context of MSD risk factor classification, aiding in the selection of the most suitable model for practical implementation.

\begin{table*}[!ht]
\centering
\caption{Evaluation metrics for the NLP models across MSD risk factor categories}
\resizebox{\linewidth}{!}{%
\begin{tabular}{|l|c|c|c|c|c|} 
\hline
\multirow{2}{*}{\textbf{Models}} & \multirow{2}{*}{\textbf{Categories}} & \multicolumn{4}{c|}{\textbf{Evaluation Metrics}} \\ 
\cline{3-6}
 &  & \textbf{Precision} & \textbf{Recall} & \textbf{F1-score} & \textbf{Accuracy} \\ 
\hline
\multirow{5}{*}{BERT+Cosine Similarity} & Biomechanical & 50.00\% & 40.00\% & 44.00\% & \multirow{5}{*}{28.00\%} \\ 
\cline{2-5}
 & Organizational & 0.00\% & 0.00\% & 0.00\% &  \\ 
\cline{2-5}
 & Personal & 0.00\% & 0.00\% & 0.00\% &  \\ 
\cline{2-5}
 & Psychosocial & 40.00\% & 80.00\% & 53.00\% &  \\ 
\cline{2-5}
 & Workplace & 25.00\% & 20.00\% & 22.00\% &  \\ 
\hline
\multirow{5}{*}{NLTK+Jaccard Similarity} & Biomechanical & 0.00\% & 0.00\% & 0.00\% & \multirow{5}{*}{20.00\%} \\ 
\cline{2-5}
 & Organizational & 0.00\% & 0.00\% & 0.00\% &  \\ 
\cline{2-5}
 & Personal & 20.00\% & 100.00\% & 33.00\% &  \\ 
\cline{2-5}
 & Psychosocial & 0.00\% & 0.00\% & 0.00\% &  \\ 
\cline{2-5}
 & Workplace & 0.00\% & 0.00\% & 0.00\% &  \\ 
\hline
\multirow{5}{*}{Sentence Transformer+Cosine Similarity} & Biomechanical & 100.00\% & 100.00\% & 100.00\% & \multirow{5}{*}{80.00\%} \\ 
\cline{2-5}
 & Organizational & 100.00\% & 100.00\% & 100.00\% &  \\ 
\cline{2-5}
 & Personal & 50.00\% & 100.00\% & 67.00\% &  \\ 
\cline{2-5}
 & Psychosocial & 100.00\% & 100.00\% & 100.00\% &  \\ 
\cline{2-5}
 & Workplace & 0.00\% & 0.00\% & 0.00\% &  \\ 
\hline
\multirow{5}{*}{Sentence Transformer+Euclidean Distance} & Biomechanical & 100.00\% & 100.00\% & 100.00\% & \multirow{5}{*}{\textbf{100.00\%}} \\ 
\cline{2-5}
 & Organizational & 100.00\% & 100.00\% & 100.00\% &  \\ 
\cline{2-5}
 & Personal & 100.00\% & 100.00\% & 100.00\% &  \\ 
\cline{2-5}
 & Psychosocial & 100.00\% & 100.00\% & 100.00\% &  \\ 
\cline{2-5}
 & Workplace & 100.00\% & 100.00\% & 100.00\% &  \\ 
\hline
\multirow{5}{*}{Sentence Transformer+Manhattan Distance} & Biomechanical & 67.00\% & 80.00\% & 73.00\% & \multirow{5}{*}{40.00\%} \\ 
\cline{2-5}
 & Organizational & 100.00\% & 20.00\% & 33.00\% &  \\ 
\cline{2-5}
 & Personal & 67.00\% & 40.00\% & 50.00\% &  \\ 
\cline{2-5}
 & Psychosocial & 25.00\% & 20.00\% & 22.00\% &  \\ 
\cline{2-5}
 & Workplace & 18.00\% & 40.00\% & 25.00\% &  \\ 
\hline
\multirow{5}{*}{Sentence Transformer+Mahalonobis Distance} & Biomechanical & 100.00\% & 100.00\% & 100.00\% & \multirow{5}{*}{80.00\%} \\ 
\cline{2-5}
 & Organizational & 100.00\% & 100.00\% & 100.00\% &  \\ 
\cline{2-5}
 & Personal & 50.00\% & 100.00\% & 67.00\% &  \\ 
\cline{2-5}
 & Psychosocial & 100.00\% & 100.00\% & 100.00\% &  \\ 
\cline{2-5}
 & Workplace & 0.00\% & 0.00\% & 0.00\% &  \\ 
\hline
\multirow{5}{*}{Sentence Transformer+Minkowski Distance} & Biomechanical & 100.00\% & 100.00\% & 100.00\% & \multirow{5}{*}{\textbf{100.00\%}} \\ 
\cline{2-5}
 & Organizational & 100.00\% & 100.00\% & 100.00\% &  \\ 
\cline{2-5}
 & Personal & 100.00\% & 100.00\% & 100.00\% &  \\ 
\cline{2-5}
 & Psychosocial & 100.00\% & 100.00\% & 100.00\% &  \\ 
\cline{2-5}
 & Workplace & 100.00\% & 100.00\% & 100.00\% &  \\ 
\hline
\multirow{5}{*}{Sentence Transformer+Bray Curtis Distance} & Biomechanical & 100.00\% & 100.00\% & 100.00\% & \multirow{5}{*}{\textbf{100.00\%}} \\ 
\cline{2-5}
 & Organizational & 100.00\% & 100.00\% & 100.00\% &  \\ 
\cline{2-5}
 & Personal & 100.00\% & 100.00\% & 100.00\% &  \\ 
\cline{2-5}
 & Psychosocial & 100.00\% & 100.00\% & 100.00\% &  \\ 
\cline{2-5}
 & Workplace & 100.00\% & 100.00\% & 100.00\% &  \\
\hline
\end{tabular}
}
\label{tab3}
\end{table*}

\begin{table*}[!ht]
\centering
\caption{Performance comparisons of the NLP models based on the macro average evaluation metrics}
\label{tab:macro_avg}
\resizebox{\linewidth}{!}{%
\begin{tabular}{| l | c | c | c |}
\hline
\textbf{Models} & \textbf{Macro Average Precision} & \textbf{Macro Average Recall} & \textbf{Macro Average F1-score} \\
\hline
BERT+Cosine Similarity & 23.00\% & 28.00\% & 24.00\% \\
\hline
NLTK+Jaccard Similarity & 4.00\% & 20.00\% & 7.00\% \\
\hline
Sentence Transformer+Cosine Similarity & 70.00\% & 80.00\% & 73.00\% \\
\hline
Sentence Transformer+Euclidean Distance & 100.00\% & 100.00\% & 100.00\% \\
\hline
Sentence Transformer+Manhattan Distance & 55.00\% & 40.00\% & 41.00\% \\
\hline
Sentence Transformer+Mahalonobis Distance & 70.00\% & 80.00\% & 73.00\% \\
\hline
Sentence Transformer+Minkowski Distance & 100.00\% & 100.00\% & 100.00\% \\
\hline
Sentence Transformer+Bray Curtis Distance & 100.00\% & 100.00\% & 100.00\% \\
\hline

\end{tabular}
}
\end{table*}

The results of this study indicate that the BERT model with cosine similarity metric achieved an overall accuracy of 28\% in predicting the appropriate label for each risk factor. The F1-score for the `psychosocial' category was 53\%, indicating that the model was relatively successful in predicting this category. However, the F1 scores for the `biomechanical,' `organizational,' `personal,' and `workplace' categories were considerably lower, indicating that the model was not as successful in predicting these categories.

Table \ref{tab:macro_avg} presents a comparative analysis of various models' performance metrics in predicting MSD risk factors. Among the models assessed, Sentence Transformer combined with Euclidean Distance achieved the highest precision, recall, and F1-score, each at 100\%. This indicates that this model consistently performed exceptionally well identifying relevant risk factors. Following closely, Sentence Transformer paired with Minkowski Distance and Bray-Curtis Distance also attained perfect scores across all metrics, suggesting robust predictive capabilities. Notably, while showing moderate performance, BERT and Cosine Similarity lagged significantly behind the top-performing models regarding precision, recall, and F1-score. Conversely, NLTK paired with Jaccard Similarity exhibited the lowest performance across all metrics, indicating its limitations in accurately identifying MSD risk factors compared to the other models evaluated.

\begin{figure*}[!ht]
    \centering
    \begin{subfigure}{0.6\linewidth}
    \includegraphics[width=1\linewidth]{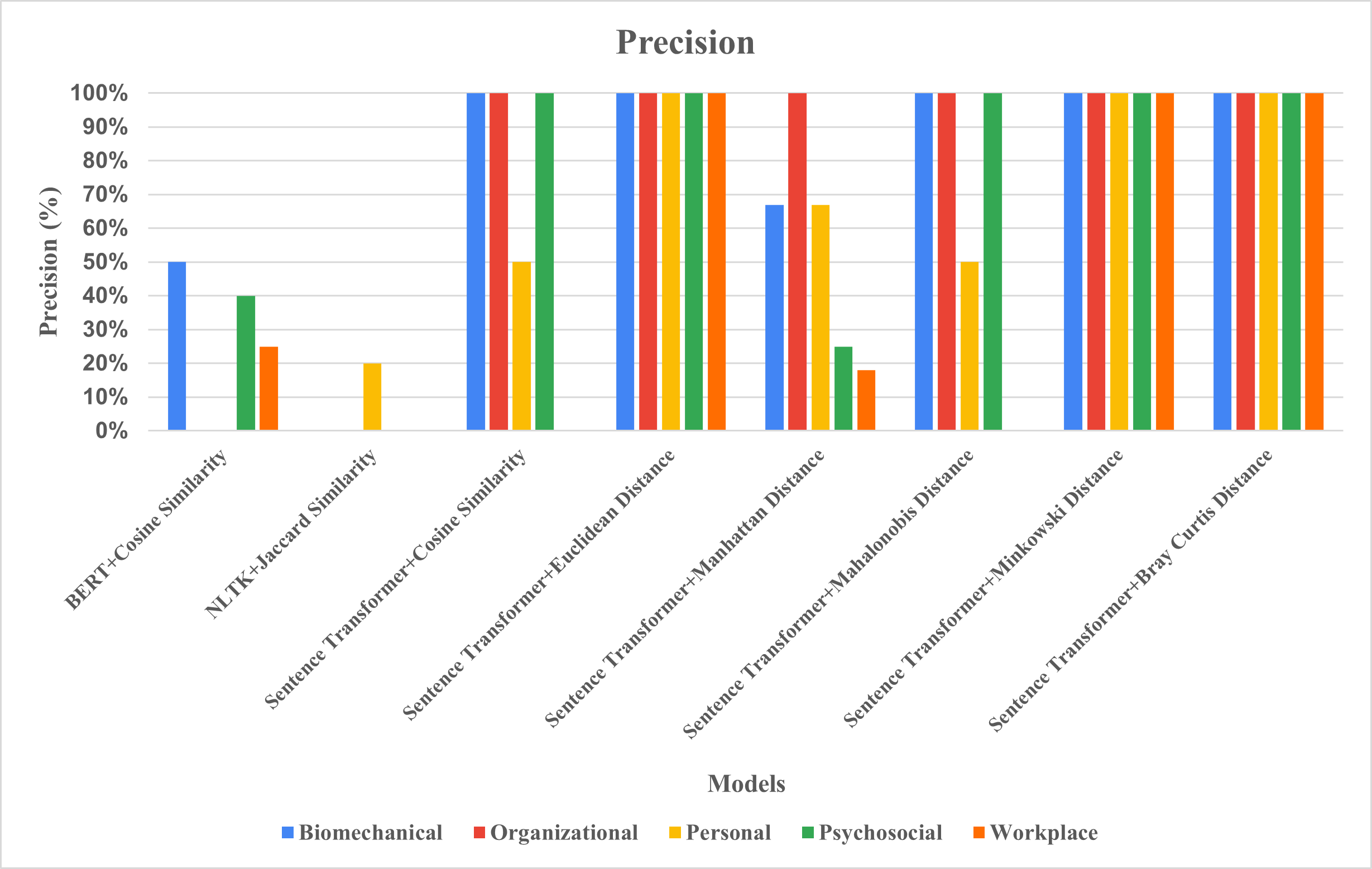}
    \caption{Precison comparison}
    \label{precision}
    \end{subfigure}
    \hfill
    \begin{subfigure}{0.6\linewidth}
    \includegraphics[width=1\linewidth]{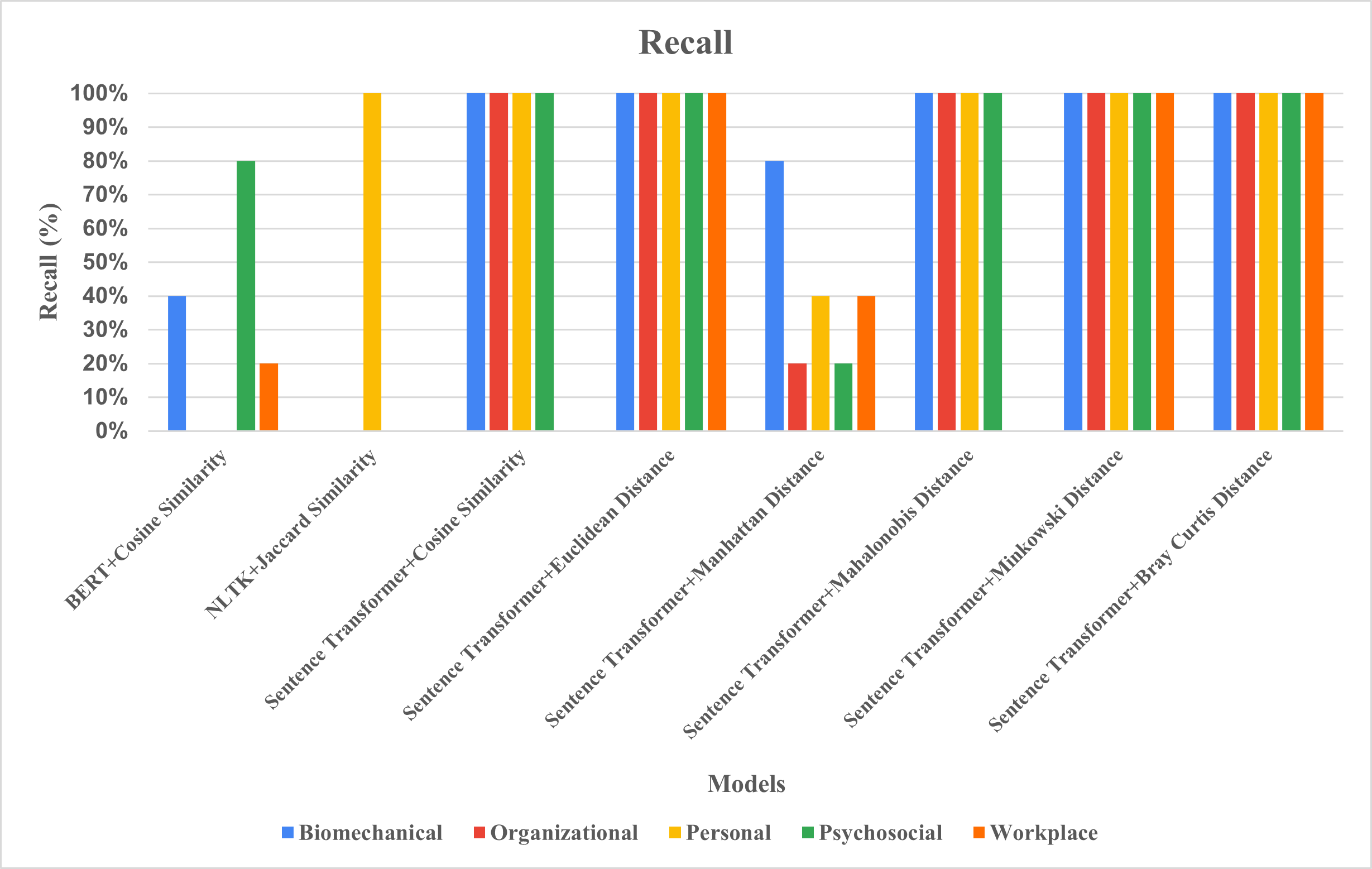}
    \caption{Recall comparison}
    \label{recall}
    \end{subfigure}
    \hfill
    \begin{subfigure}{0.6\linewidth}
    \includegraphics[width=1\linewidth]{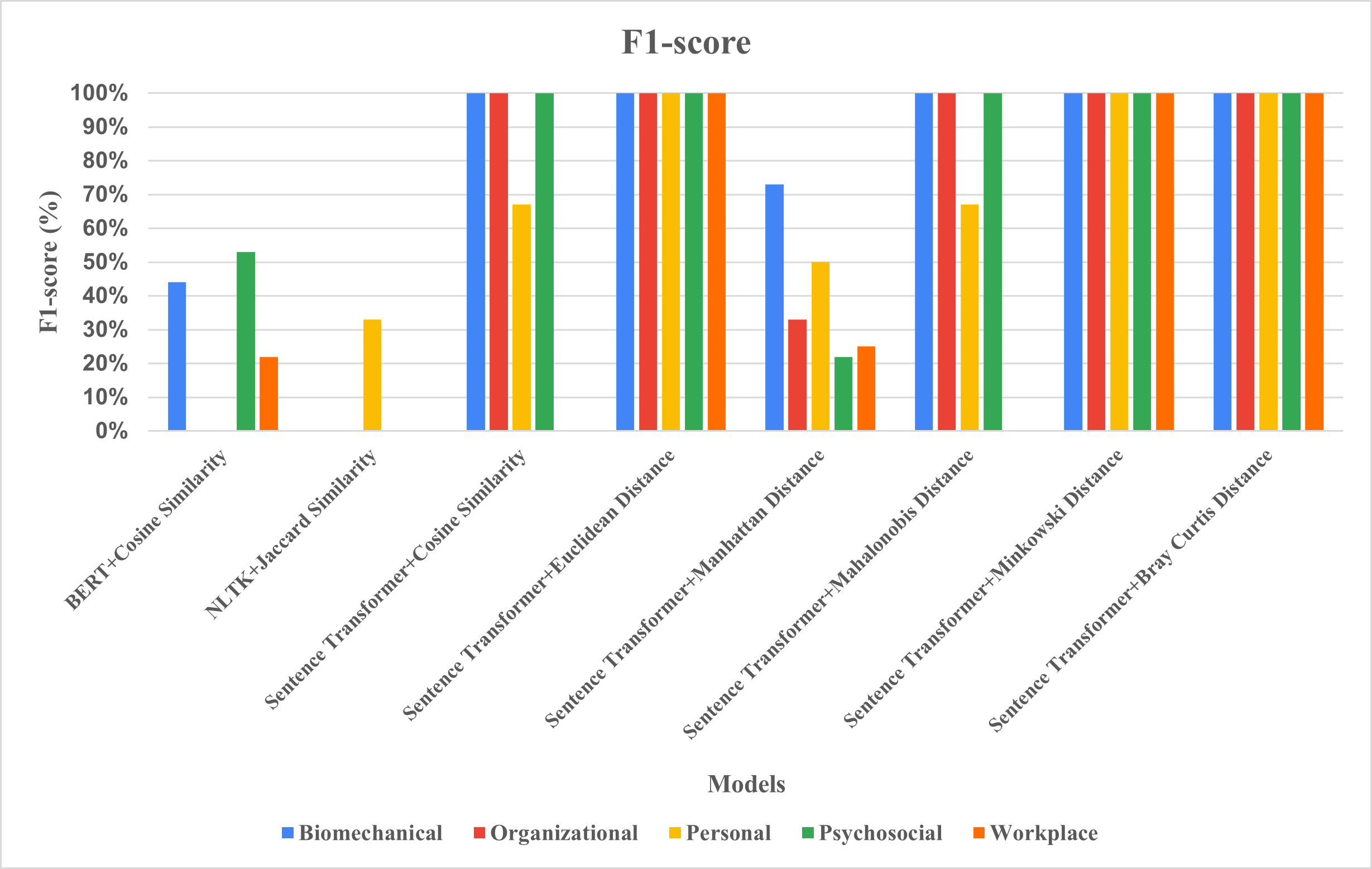}
    \caption{F1-score comparison}
    \label{f1-score}
    \end{subfigure}
\caption{Precision, Recall, and F1-score comparison among the implemented NLP models.}
\label{fig:comparison_pr_re_f1}
\end{figure*}

We can observe distinct performance trends across the implemented models based on the precision, recall, and F1-scores depicted in Figure \ref{fig:comparison_pr_re_f1}. The precision graph highlights the ability of each model to classify positive instances among all instances predicted as positive correctly. In contrast, the recall graph illustrates the models' effectiveness in capturing all positive instances from the dataset. Additionally, the F1-score graph provides a balanced measure of a model's precision and recall, offering insights into its overall performance.

Analyzing Figure \ref{precision}, Sentence Transformer with Euclidean, Minkowski, or Bray-Curtis distance metrics demonstrates the highest precision, indicating its proficiency in correctly identifying positive instances with minimal false positives. However, Sentence Transformer with Mahalonobis distance and Cosine Similarity exhibits slightly lower precision but compensates with higher recall in categorizing personal risk factors, suggesting a trade-off between precision and recall. Conversely, NLTK with Jaccard Similarity displays the lowest precision among the models, indicating a higher rate of false positives.

Regarding Figure \ref{recall}, the Sentence Transformer with Euclidean, Minkowski, or Bray-Curtis distance metrics outperforms the other models by capturing a larger proportion of positive instances from the dataset. BERT+Cosine Similarity shows low precision and demonstrates low recall, indicating potential limitations in identifying positive instances effectively.

Considering the Figure \ref{f1-score}, Sentence Transformer with Euclidean, Minkowski, or Bray-Curtis distance metrics achieves the highest balance between precision and recall, reflecting its overall robustness in classification performance. Sentence Transformer with Cosine Similarity or Mahalonobis or Manhattan distance metrics follows closely behind, exhibiting a slightly lower F1-score but maintaining a strong balance between precision and recall. On the other hand, NLTK+Jaccard Similarity and BERT+Cosine Similarity lag behind the other models, indicating a suboptimal balance between precision and recall, which may affect its overall effectiveness in classification tasks.

The use of eight distinct NLP models for risk factor classification, each employing different similarity or distance measures, introduces a rich landscape of accuracies with associated strengths and limitations. Starting with BERT coupled with the cosine similarity measure, its strength lies in its ability to capture contextual relationships, offering a nuanced understanding of risk factors. However, its computational intensity might hinder real-time applications. The cosine similarity measure, on its own, is computationally efficient but may struggle with capturing complex contextual nuances. Moving to distance-based measures, the Manhattan distance method is effective in capturing differences in individual risk factors but may oversimplify the overall relationships. NLTK, coupled with the Jaccard similarity measure, brings linguistic analysis into play, offering interpretability, yet it may struggle to capture the semantic richness present in the MSD risk factors. Euclidean distance, Mahalanobis distance, Bray-Curtis distance, and Minkowski distance measures provide varying degrees of sensitivity to feature space, offering a trade-off between computational efficiency and nuanced analysis.

The implications of varying accuracies among these models are multifold. On one hand, the diversity of models allows for a comprehensive understanding of risk factors from different perspectives. On the other hand, the discrepancies in accuracies raise questions about the robustness of certain models in specific contexts. It is essential to weigh the computational demands against the interpretability and nuanced understanding provided by each model. The choice of a model should align with the specific goals of the research, considering factors such as real-time applicability, interpretability, and the desired level of granularity in risk factor analysis. Additionally, the varying accuracies highlight the complexity of MSD risk factor classification, urging researchers to explore ensemble methods or hybrid approaches to leverage the strengths of multiple models and mitigate their individual limitations.

The sentence transformer, along with Euclidean, Bray-Curtis, and Minkowski distance measures, resulted in a perfect accuracy score of 100\%. The precision, recall, and f1-score were also 100\% for each class, including biomechanical, organizational, personal, psychosocial, and workplace. Next, the Mahalanobis distance measure was implemented, and the accuracy score was 80\%. The precision, recall, and f1-score for biomechanical, organizational, and psychosocial classes were 100\%, while for the personal and workplace classes, the precision, recall, and f1-score were 50\% and 0\%, respectively. The four distance measures showed promising results in classifying musculoskeletal risk factors. The Euclidean, Minkowski, and Bray-Curtis measures yielded perfect classification results, while the Mahalanobis measure performed similarly well, with a few misclassifications in some classes. 

Overall, these results indicate that the sentence transformer combined with specific distance measures, such as Euclidean, Bray-Curtis, and Minkowski, demonstrated robust performance in MSD risk factor classification. The Mahalanobis distance measure also showed promise, with minor misclassifications in certain classes.

\subsection{Statistical test}

\begin{table*}[!ht]
\centering
\caption{Paired t-test results and Cohen's d values for model performance comparison}
\resizebox{\linewidth}{!}{
\label{tab:statistical-test}
\begin{tabular}{|l |l |l |l |l|} \hline  
\textbf{Model 1} & \textbf{Model 2} & \textbf{t-statistic} & \textbf{p-value} & \textbf{Cohen's d} \\ \hline 
BERT+Cosine Similarity & NLTK+Jaccard Similarity & 11.75894 & 9.15E-07 & 4.308289 \\ \hline 
BERT+Cosine Similarity & Sentence Transformer+Cosine Similarity & -103.524 & 3.72E-15 & -25.6322 \\ \hline 
BERT+Cosine Similarity & Sentence Transformer+Euclidean Distance & -101 & 4.64E-15 & -43.071 \\ \hline 
BERT+Cosine Similarity & Sentence Transformer+Manhattan Distance & -14.9558 & 1.16E-07 & -5.94054 \\ \hline 
BERT+Cosine Similarity & Sentence Transformer+Mahalanobis Distance & -60.6038 & 4.57E-13 & -24.214 \\ \hline 
BERT+Cosine Similarity & Sentence Transformer+Minkowski Distance & -87.75 & 1.64E-14 & -39.449 \\ \hline 
BERT+Cosine Similarity & Sentence Transformer+Bray Curtis Distance & -84.6313 & 2.27E-14 & -37.8482 \\ \hline 
NLTK+Jaccard Similarity & Sentence Transformer+Cosine Similarity & -92.2987 & 1.04E-14 & -35.1451 \\ \hline 
NLTK+Jaccard Similarity & Sentence Transformer+Euclidean Distance & -229.824 & 2.84E-18 & -63.7418 \\ \hline 
NLTK+Jaccard Similarity & Sentence Transformer+Manhattan Distance & -22.6947 & 2.97E-09 & -10.9457 \\ \hline 
NLTK+Jaccard Similarity & Sentence Transformer+Mahalanobis Distance & -89.4554 & 1.38E-14 & -31.8249 \\ \hline 
NLTK+Jaccard Similarity & Sentence Transformer+Minkowski Distance & -117 & 1.24E-15 & -55.3082 \\ \hline 
NLTK+Jaccard Similarity & Sentence Transformer+Bray Curtis Distance & -110.573 & 2.05E-15 & -51.8114 \\ \hline 
Sentence Transformer+Cosine Similarity & Sentence Transformer+Euclidean Distance & -31.3793 & 1.66E-10 & -12.9493 \\ \hline 
Sentence Transformer+Cosine Similarity & Sentence Transformer+Manhattan Distance & 41.65122 & 1.32E-11 & 18.38694 \\ \hline 
Sentence Transformer+Cosine Similarity & Sentence Transformer+Mahalanobis Distance & -2.91318 & 0.017222 & -1.0903 \\ \hline 
Sentence Transformer+Cosine Similarity & Sentence Transformer+Minkowski Distance & -20.4465 & 7.47E-09 & -11.4668 \\ \hline 
Sentence Transformer+Cosine Similarity & Sentence Transformer+Bray Curtis Distance & -30.0416 & 2.45E-10 & -10.9697 \\ \hline 
Sentence Transformer+Euclidean Distance & Sentence Transformer+Manhattan Distance & 69.46264 & 1.34E-13 & 33.15995 \\ \hline 
Sentence Transformer+Euclidean Distance & Sentence Transformer+Mahalanobis Distance & 23.19866 & 2.44E-09 & 9.651796 \\ \hline 
Sentence Transformer+Euclidean Distance & Sentence Transformer+Minkowski Distance & 0.807573 & 0.440158 & 0.421741 \\ \hline 
Sentence Transformer+Euclidean Distance & Sentence Transformer+Bray Curtis Distance & 0.646997 & 0.533786 & 0.307794 \\ \hline 
Sentence Transformer+Manhattan Distance & Sentence Transformer+Mahalanobis Distance & -38.0625 & 2.96E-11 & -17.7475 \\ \hline 
Sentence Transformer+Manhattan Distance & Sentence Transformer+Minkowski Distance & -82.1815 & 2.96E-14 & -30.5747 \\ \hline 
Sentence Transformer+Manhattan Distance & Sentence Transformer+Bray Curtis Distance & -58.5834 & 6.2E-13 & -29.459 \\ \hline 
Sentence Transformer+Mahalanobis Distance & Sentence Transformer+Minkowski Distance & -19.3261 & 1.23E-08 & -8.7355 \\ \hline 
Sentence Transformer+Mahalanobis Distance & Sentence Transformer+Bray Curtis Distance & -27.1307 & 6.08E-10 & -8.4624 \\ \hline 
Sentence Transformer+Minkowski Distance & Sentence Transformer+Bray Curtis Distance & -0.12577 & 0.902681 & -0.06802 \\ \hline

\end{tabular}
}
\end{table*}

The statistical analysis conducted in this study followed a rigorous methodology that began with a 10-fold cross-validation in which the dataset was divided into ten subsets, ensuring each subset represented an equal proportion of the data and maintaining the integrity of the original dataset's distribution. Each model was then trained and evaluated ten times, with a different subset serving as the testing set in each iteration, while the remaining nine subsets were used for training. After completing the 10-fold cross-validation procedure, statistical tests were conducted on the accuracies obtained from each model. The paired t-tests were conducted to compare the accuracies of different models in terms of their t-statistic, p-value, and Cohen's d-values. The null hypothesis $(H_0)$ for each comparison was that there was no significant difference in accuracies between the two models being compared. The alternative hypothesis $(H_1)$ was that the two models had a significant difference in accuracies. The decisions for each comparison were made based on the p-value and Cohen's d-values. If the p-value was below the significance level (e.g., 0.05) and the absolute value of Cohen's d exceeded a threshold (e.g., 0.5), the $H_0$ was rejected, and practical significance was considered. $H_0$ was rejected if only the p-value were below the significance level. Practical significance was considered if only the absolute value of Cohen's d exceeded the threshold. No significant difference or practical significance was concluded if neither criterion was met.

comparisons between BERT+Cosine Similarity and NLTK+Jaccard Similarity, as well as between various Sentence Transformer models, consistently rejected the $H_0$, indicating substantial disparities in performance (see Table \ref{tab:statistical-test}). Additionally, these differences were statistically significant and practically meaningful, as indicated by the consideration of Cohen's d effect size. Specifically, Sentence Transformer models consistently outperformed other models across different similarity metrics, suggesting their potential superiority in the evaluated task. These findings underscore the importance of considering statistical and practical significance when interpreting model performance comparisons and informing decision-making processes in model selection and deployment for optimal outcomes.

\subsection{Ranking of risk factors and implications for preventive measures}

Table \ref{tab1} presents the ranking results for MSD risk factors based on 1050 survey participant responses. The participants ranked each MSD risk factor on a scale of 1 to 25 based on their severity. Several noteworthy observations emerge from the ranking analysis.

Firstly, ``Working posture" consistently emerged as the most severe risk factor, occupying the top spot with a ranking of 1. This underscores the widely recognized importance of maintaining proper posture in mitigating MSD. ``Repetitive motion" is closely followed as the second most severe risk factor, reaffirming the significance of addressing repetitive tasks in workplace ergonomics. Surprisingly, ``Layout" secured the third position, shedding light on the substantial role that the physical arrangement of workspaces plays in shaping MSD risk perceptions. Conversely, ``Job insecurity" was ranked as the least severe risk factor, obtaining the highest ranking of 25. This suggests that participants perceived job security as less directly associated with MSD problems than other factors. Similarly, ``Effort reward imbalance" and ``Poor employee facility" received notably high rankings, indicating that participants considered these factors as critical contributors to MSD risks. The rankings revealed gender-based differences in MSD risk perception, with ``Gender" receiving a relatively lower ranking (16) among the factors. This result implies that respondents did not consider gender as a predominant factor in MSD susceptibility.

The analysis of job insecurity as an MSD risk factor revealed compelling associations with physical health outcomes. Individuals experiencing job insecurity exhibited a higher prevalence of MSD symptoms. The stress and anxiety associated with uncertain employment conditions can manifest physically, contributing to the development or exacerbation of musculoskeletal issues. This finding underscores the importance of considering psychosocial factors, such as job insecurity, in comprehensive approaches to prevent and manage MSD. The exploration of the Effort Reward Imbalance in the context of MSD provided nuanced insights into the intricate relationship between occupational effort, perceived rewards, and musculoskeletal health. Employees experiencing an imbalance between their efforts exerted at work and the rewards received were found to have an elevated risk of developing MSD. This suggests that beyond physical strain, the psychosocial aspect of work, including perceptions of fairness and recognition, significantly contributes to the musculoskeletal well-being of workers. Addressing ERI could be integral to holistic interventions aiming to reduce MSD prevalence. Examining the impact of inadequate employee facilities on MSD risk highlighted the significance of the physical work environment. Employees reporting poor facilities, such as uncomfortable seating, lack of ergonomic equipment, or insufficient workspace, exhibited a higher likelihood of experiencing musculoskeletal discomfort. This underscores the importance of optimizing workplace ergonomics and providing suitable infrastructure to mitigate the physical strain associated with suboptimal working conditions.

The identification of severe risk factors, particularly ``Working Posture," carries immense significance in the context of occupational health and has profound implications for preventive measures. Poor working posture has been consistently associated with musculoskeletal discomfort and disorders, making it a critical focal point for intervention. The significance lies in the potential for adverse health outcomes, including chronic pain, reduced productivity, and increased absenteeism among workers. Addressing the issue of working posture is crucial for several reasons. Firstly, prolonged periods of poor posture can lead to musculoskeletal imbalances, contributing to the development of conditions such as lower back pain, neck strain, and repetitive strain injuries. These conditions not only adversely affect individual well-being but can also result in increased healthcare costs and reduced overall workplace productivity. The identified severe risk of working posture also underscores the importance of ergonomic interventions in the workplace. Implementing ergonomic assessments and interventions, such as adjustable furniture, proper chair design, and regular breaks to encourage movement, can significantly mitigate the impact of poor working posture. Training programs on ergonomics and postural awareness can empower employees to adopt healthier work habits, preventing the onset of musculoskeletal issues. From a preventive standpoint, organizations should prioritize the design of workspaces that promote optimal ergonomic conditions. This includes providing ergonomic furniture, ensuring proper workstation setups, and offering education on maintaining good posture. Integrating regular breaks and incorporating exercises that target postural muscles can further contribute to preventing the development of MSDs.

Notably, the rankings produced by our mode-based measurement of survey data align precisely with the rankings derived from the previous literature by \cite{talapatra_musculoskeletal_2022}, which were generated using the Fuzzy Analytic Hierarchy Process (Fuzzy AHP). This convergence of rankings reinforces the robustness and reliability of our mode-based approach in assessing the perceived severity of MSD risk factors, corroborating the findings of prior research conducted by \cite{talapatra_musculoskeletal_2022}.

The mode-based ranking approach employed here provides insights into the perceived severity of various MSD risk factors, emphasizing the importance of addressing risk factors that are collectively regarded as more severe. These findings can guide targeted interventions and preventive measures aimed at reducing the prevalence of MSD in occupational settings.

\subsection{Managerial implications}
The results of our study provide insightful information with important management ramifications for companies that prioritize worker well-being and occupational health. An intricate comprehension of the risk factors for MSD has been made possible by integrating mode-based ranking and NLP-based classification approaches. With a thorough grasp of MSD risk variables provided by the study, managers are better equipped to allocate resources and decide on intervention tactics and preventative measures. Managers may optimize resource usage and enhance overall organizational efficiency by prioritizing efforts that target the most important risk factors for MSDs by applying the information acquired from the study. Knowing the mode-based ranking and classification of MSD risk factors allows managers to create customized intervention methods suited to their workforce's unique demands and difficulties. For instance, managers can implement targeted interventions, like ergonomic assessments, workplace modifications, or employee training programs, to mitigate risks and promote the health and well-being of employees if certain risk factors are found to be particularly prevalent or severe within the organization. Managers can use the comparison between NLP approaches and distance measurements as a benchmark to assess how well their organization's current classification models and procedures are working. Managers may evaluate the efficacy of their present strategies and pinpoint areas for improvement by benchmarking against industry standards and best practices highlighted in the research. As a result, performance and results are improved via ongoing learning and process improvement inside the business. The results of this study can help healthcare companies with their long-term strategic planning initiatives.  Managers can use the insights gleaned from the research to develop strategic initiatives to proactively address MSD risk factors, reduce healthcare costs, and improve employee productivity and satisfaction. By aligning organizational goals and priorities with the study's findings, managers can ensure that resources are allocated effectively and strategic objectives are achieved.

\subsection{Practical applications for organizations and occupational health practitioners}
The practical applications of the study's findings hold significant implications for organizations and occupational health practitioners aiming to foster healthier and more productive workplaces. Firstly, acknowledging the impact of psychosocial factors such as job insecurity and effort-reward imbalance on MSD underscores the need for organizations to prioritize employee well-being beyond physical conditions. Implementing strategies to reduce job insecurity, such as transparent communication and employee engagement initiatives, can contribute to a healthier work environment. Moreover, addressing effort-reward imbalances by recognizing and rewarding employees' contributions fosters a positive organizational culture. Additionally, identifying poor employee facilities as a significant risk factor emphasizes investing in ergonomic workspaces and ensuring employees have access to adequate facilities. Occupational health practitioners can leverage these findings to tailor preventive interventions, conduct targeted ergonomic assessments, and provide employee education on mitigating the identified risk factors. Integrating these insights into organizational policies and practices can ultimately enhance overall workplace health, reduce the prevalence of MSD, and contribute to a more sustainable and supportive work environment.

\section{Conclusions and future research directions}
\label{sec:conclusion}
This research has comprehensively explored the risk factors of MSD, employing an innovative approach that integrates NLP techniques and mode-based ranking methodologies. The study aimed to advance the understanding of MSD risk factors, their classification, and their relative severity, thereby contributing to enhanced preventive and management strategies. The research holds broad implications for the field of occupational health, offering valuable insights that can shape both research endeavors and practical interventions. Firstly, the study underscores the need for a holistic approach to occupational health beyond traditional physical risk factors. Integrating psychosocial elements such as job insecurity and effort-reward imbalance into occupational health frameworks acknowledges the interconnected nature of mental and physical well-being in the workplace. Moreover, using advanced NLP pre-trained transformers in risk factor analysis presents a paradigm shift in occupational health research. The study demonstrates the applicability of NLP not only in linguistic analysis but also in the systematic classification and understanding of complex risk factors. This suggests a broader potential for NLP in processing and extracting valuable insights from large volumes of textual data in the field. From a practical standpoint, identifying specific risk factors like inadequate employee facilities offers actionable points for intervention. Organizations can use these insights to redesign workspaces, implement ergonomic improvements, and invest in employee well-being programs. Additionally, the emphasis on the psychosocial aspects of work highlights the importance of organizational culture, leadership, and communication in promoting a healthy workplace.

The classification of 25 MSD risk factors was executed through eight distinct models, leveraging pre-trained transformers, cosine similarity, and various distance metrics. The comprehensive evaluation, as illustrated in Table \ref{tab3}, elucidates the strengths and weaknesses of each model in categorizing risk factors into personal, biomechanical, workplace, psychological, and organizational classes. Notably, the BERT model with cosine similarity demonstrated an overall accuracy of 28\%, while the sentence transformer and Euclidean, Bray-Curtis, and Mahalonobis distances achieved a perfect accuracy score of 100\%. The sentence transformer with cosine similarity or Mahalonobis distance achieved 80\% accuracy.

The mode-based ranking approach applied to survey data, illustrated in Table \ref{tab1}, yielded intriguing results. ``Working posture" emerged as the most severe risk factor, emphasizing the paramount importance of maintaining proper posture in preventing MSD. ``Repetitive motion" closely followed, underscoring the significance of addressing repetitive tasks in ergonomic considerations. The rankings aligned precisely with the previous literature of \cite{talapatra_musculoskeletal_2022}, further validating our approach. Additionally, the survey participants collectively perceived ``Job insecurity," ``Effort reward imbalance," and ``Poor employee facility" as critical contributors to MSD risks.

The convergence of rankings between our mode-based approach and Fuzzy AHP approach by \cite{talapatra_musculoskeletal_2022} rankings emphasizes the consistency and reliability of our findings. These results are valuable for targeted interventions and preventive measures in occupational settings. Prioritizing risk factors based on collective perceptions can guide effective strategies for reducing the prevalence of MSD. It is recommended that organizations focus on improving workplace conditions, addressing job insecurity concerns, and enhancing employee facilities to mitigate the impact of MSD risk factors.

While this study provides valuable insights, it is not without limitations. The reliance on survey data introduces subjective elements, and the demographic characteristics of participants may influence the generalizability of findings. One notable limitation is the sample size, as a larger and more diverse dataset could provide a more comprehensive representation of the population. Additionally, the demographics of the surveyed individuals might not fully capture the heterogeneity present in various occupational settings, potentially introducing biases in our analysis. Moreover, the voluntary nature of survey participation may lead to a self-selection bias, as those experiencing MSD issues may be more inclined to respond. To enhance the transparency and reliability of our results, we recognize the need for caution in extrapolating findings to broader populations and emphasize the importance of future research with larger, more diverse samples to validate and extend our insights into MSD risk factors.

Several promising avenues for future investigations can be explored to advance our understanding of MSD risk factors and further enhance the application of NLP in occupational health research. First, expanding the survey scope to include a more diverse and representative sample from various occupational sectors and demographics would contribute to a richer dataset, allowing for a more nuanced analysis of MSD risk factors across different contexts. Furthermore, future research could develop hybrid models that leverage the strengths of linguistic analysis models like GPT-2, GPT-3, RoBERTa, XLNet, and XLM, as well as explore other distance-based measures. This hybrid approach may provide a more comprehensive understanding of the interplay between linguistic nuances and quantitative relationships within MSD risk factors. Additionally, longitudinal studies could be conducted to track changes in MSD risk factors over time, enabling a dynamic analysis of how occupational health conditions evolve. Incorporating real-time monitoring and wearable sensor data could offer valuable insights into the daily activities and environmental factors contributing to MSD. Exploring the integration of explainable AI techniques would enhance the interpretability of NLP models, addressing concerns related to the ``black-box" nature of some advanced algorithms. This could be particularly relevant in occupational health settings where transparent decision-making processes are crucial. Lastly, collaborative efforts between researchers, healthcare professionals, and industry stakeholders could facilitate the development of practical interventions and preventive measures based on the identified risk factors. Implementing and evaluating these interventions in real-world occupational settings would contribute to translating research findings into tangible improvements in workplace health and safety.

This research offers a holistic approach to MSD risk factor analysis, leveraging NLP and mode-based ranking to provide nuanced insights for prevention and management strategies. The findings contribute to the evolving landscape of occupational health, urging continued exploration and innovation in addressing the multifaceted challenges of MSD.

%\bibliography{sample}

\bibliographystyle{apalike} 
\bibliography{references}

\end{document}